\documentclass[pdflatex]{sn-jnl}

\usepackage{graphicx}%
\usepackage{multirow}%
\usepackage{amsmath,amssymb,amsfonts}%
\usepackage{amsthm}%
\usepackage{mathrsfs}%
\usepackage{xcolor}%
\usepackage{textcomp}%
\usepackage{manyfoot}%
\usepackage{rotating}  
\usepackage{booktabs}%
\usepackage{algorithm}%
\usepackage{algorithmicx}%
\usepackage{algpseudocode}%
\usepackage{listings}%
\usepackage[
  backend=biber,
  style=apa,
  sortcites=true,
  sorting=nyt,
  url=false,
  doi=true,
  eprint=false
]{biblatex}
\usepackage{amsmath}
\usepackage{amssymb}
\usepackage{booktabs}
\usepackage{tabularx}
\usepackage{float}
\usepackage{multirow}
\usepackage{siunitx}
\usepackage{pdflscape} 
\usepackage{makecell}
\usepackage{lineno}
\usepackage{caption}
\usepackage{algorithm}
\usepackage{geometry}
\usepackage{xcolor}
\usepackage{tcolorbox}
\usepackage{algpseudocode}
\usepackage{color}
\usepackage{hyperref}
\usepackage{graphicx} 
\usepackage{epsfig}
\usepackage{epstopdf} 
\usepackage{setspace} 
\makeatletter
\newenvironment{breakablealgorithm}
  {
   \begin{center}
     \refstepcounter{algorithm}
     \hrule height.8pt depth0pt \kern2pt
     \renewcommand{\caption}[2][\relax]{
       {\raggedright\textbf{\ALG@name~\thealgorithm} ##2\par}%
       \ifx\relax##1\relax 
         \addcontentsline{loa}{algorithm}{\protect\numberline{\thealgorithm}##2}%
       \else 
         \addcontentsline{loa}{algorithm}{\protect\numberline{\thealgorithm}##1}%
       \fi
       \kern2pt\hrule\kern2pt
     }
  }{
     \kern2pt\hrule\relax
   \end{center}
  }
\makeatother

\begin{document}
\theoremstyle{thmstyleone}%
\newtheorem{theorem}{Theorem}
\newtheorem{proposition}[theorem]{Proposition}%

\theoremstyle{thmstyletwo}%
\newtheorem{example}{Example}%
\newtheorem{remark}{Remark}%

\theoremstyle{thmstylethree}%
\newtheorem{definition}{Definition}%

\raggedbottom


\title[Embark Now: User Demand Oriented Framework for Multi-day Urban Travel Itinerary Planning]{Embark Now: User Demand Oriented Framework for Multi-day Urban Travel Itinerary Planning}

\author[1]{\fnm{Rongbo} \sur{Qi}}\email{rongbo.qi@mail.nankai.edu.cn}

\author[1]{\fnm{Yaqi} \sur{Zhang}}\email{yaqi.zhang@mail.nankai.edu.cn}

\author[2]{\fnm{Shijun} \sur{Yan}}\email{shijunyan@jxnu.edu.cn}

\author[3]{\fnm{Xuemeng} \sur{Liu}}\email{xuemeng.liu@mail.nankai.edu.cn}

\author[1]{\fnm{Xiangrui} \sur{Cai}}\email{caixr@nankai.edu.cn}

\author*[1]{\fnm{Chunyao} \sur{Song}}\email{chunyao.song@nankai.edu.cn}

\affil*[1]{\orgdiv{College of Computer Science}, \orgname{Nankai University}, \orgaddress{\city{Tianjin}, \postcode{300350}, \state{Tianjin}, \country{China}}}

\affil[2]{\orgname{Jiangxi Normal University}, \orgaddress{\city{Nanchang}, \postcode{330022}, \state{Jiangxi}, \country{China}}}

\affil[3]{\orgdiv{College of Software}, \orgname{Nankai University}, \orgaddress{\city{Tianjin}, \postcode{300350}, \state{Tianjin}, \country{China}}}












\abstract{In large urban areas, planning multi-day travel itineraries is challenging due to the abundance of Points of Interest (POIs), diverse user preferences, and constraints such as opening hours. Effective solutions must dynamically accommodate diverse traveler requirements while optimizing for satisfaction and feasibility within limited computation time.
This paper addresses these challenges through introducing an innovative framework that integrates Large Language Models (LLMs) to dynamically capture user requirements with precision and flexibility, and an enhanced Greedy Randomized Adaptive Search Procedure (GRASP) algorithm as a well-suited preference-aware planner to generate feasible multi-day itineraries.
The effectiveness of our 
integrated approach is demonstrated through extensive experiments on two real-world urban datasets from Beijing and Tianjin. 
Our framework significantly outperforms state-of-the-art (SOTA) methods, improving the average total itinerary score by at least 4.52\% and 11.09\% across 5,040 user cases with diverse preferences in the two datasets. Furthermore, through end-to-end algorithmic enhancements, it achieves notable average improvements of 17.95\% and 26.07\% in the computed metrics, while also delivering substantial gains in time efficiency -- realizing average performance increases of 4.64\% and 25.55\% within shorter computation times compared to suboptimal methods that require multiple iterations.
These outcomes underscore our method's superiority in delivering both enhanced itinerary quality and computational efficiency over existing methodologies.
}




\maketitle

\section{Introduction}
With advancements in transportation and tourism, many travelers are inclined to explore unfamiliar cities
\parencite{halder2024survey}.
Despite the abundance of detailed POI (Points of Interest) data offered by various travel assistant apps, including attractions, dining options, and accommodations, as well as the curated travel itineraries provided by professional travel agencies, users still confront significant challenges when it comes to swiftly crafting a comprehensive and personalized multi-day travel plan in a large city for themselves or their families in an unfamiliar city \parencite{DBLP:journals/ipm/YheeSLK23, DBLP:journals/ipm/OrabiAA25}. These challenges stem not only from the large solution space, but also from diverse user preferences (e.g., fast-paced vs.\ slow-paced), and from spatiotemporal constraints (e.g., opening hours, travel times, and daily time budgets). We refer to this setting as the Multi-day Urban Travel Itinerary Planning (MUTIP) problem. 
Hence, the development of automated algorithms that are capable of accurately discerning diverse and user-specific preferences becomes imperative for crafting tailored itineraries that cater to individual needs.

Conventionally, this obstacle is studied under 
the Tourist Trip Design Problem (TTDP)~\parencite{Vansteenwegen_Souffriau_Oudheusden_2011,Liao_Zheng_2018}, a complex issue that is often decomposed into two distinct yet interconnected components: the Orienteering Problem (OP)~\parencite{urrutia2021variable,verbeeck2016solving} and the Team Orienteering Problem (TOP)~\parencite{bianchessi2018branch,Hu_Fathi_Pardalos_2018}. Both of these problems have garnered significant attention and have been the subject of exhaustive research endeavors over the past decade~\parencite{Gavalas_Konstantopoulos_Mastakas_Pantziou_2014,paulavivcius2023novel,kolaee2024sustainable}. 
These algorithms generally involve designing optimization objectives that cater to 
user or system requirements 
and seek locally optimal solutions using integer programming or metaheuristic approaches~\parencite{exposito2019fuzzy,tlili2021simulated}. However, their application in user-centric intelligent systems remains scarce, primarily hindered by two notable constraints. 
Firstly, their iterative essence, while aimed at identifying locally optimal solutions, can significantly prolong the process, ultimately detracting from the overall user experience due to excessive wait times. Secondly, the prerequisite of manually defining a myriad of constraints necessitates extensive learning and setup, placing undue temporal and cognitive strains on users, thereby limiting their practical adoption. 

In recent years, large language models (LLMs), such as GPT-5~\parencite{openai2025introducinggpt5} and DeepSeek-R1 \parencite{deepseekai2025deepseekr1incentivizingreasoningcapability},
have made significant strides, showing great promise in understanding user needs and reasoning. Consequently, many recent efforts have focused on employing LLMs for planning tasks~\parencite{kambhampati2024llms,hao2024large,gundawar2024robust}. 
However, LLM-only planning often struggles to consistently satisfy concrete spatiotemporal constraints and to avoid hallucinated or infeasible POIs, especially in multi-day settings~\parencite{valmeekam2024llmscantplanlrms}.

\begin{figure}[h]\centering
    \makebox[\columnwidth][c]{\includegraphics[width=1.00\columnwidth]{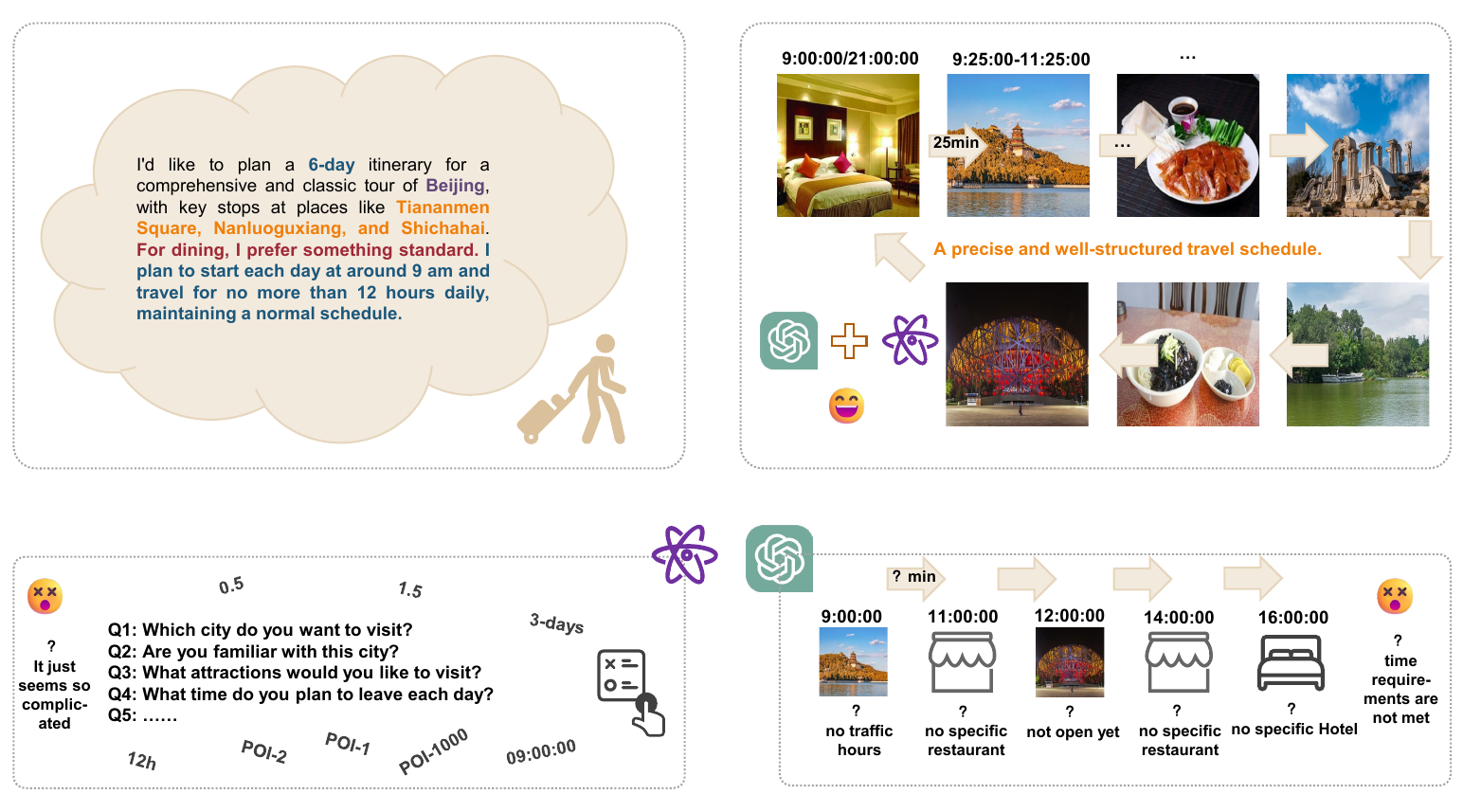}}
    \caption{Challenges and Solutions: Hybrid Travel Itinerary Planning}
    \label{fig:intro1}
\end{figure}

To further clarify the preceding problems, 
Fig.~\ref{fig:intro1} illustrates the challenges faced when relying solely on traditional methods or LLMs in isolation. 
Traditional optimization methods can produce controllable solutions under explicit constraints, but the interaction required to specify preferences and parameters is often cumbersome for users. In contrast, LLMs enable natural-language interaction, but their outputs may miss essential constraints and feasibility checks. This has led to hybrid approaches that combine LLM-based preference understanding with constraint-aware planners~\parencite{de2024trip,tang2024synergizing}.
As such, hybrid approaches emerge as the promising direction 
striking a balance between the two extremes. Such methodologies leverage the strengths of both traditional methods and LLMs, addressing the limitations of each individual approach while enhancing the overall user experience and satisfaction. These hybrid methods~\parencite{de2024trip,tang2024synergizing} initially tap into the capabilities of LLMs to profoundly understand user preferences, subsequently employing traditional planning algorithms to generate outcomes that are not only precise but also align more closely with users' expectations. While recent hybrid approaches integrate LLMs with solvers (e.g., SMT~\parencite{hao2024large}), they often struggle with the trade-off between \textbf{computational efficiency and itinerary rationality in multi-day scenarios}. General-purpose solvers can become computationally prohibitive when handling intricate spatiotemporal constraints over multiple days. Moreover, existing methods often overlook the necessity of the importance of understanding and satisfying users' routine requirements, such as prioritizing points of interest, adhering to time preferences, and catering to dining habits. Consequently, the itinerary planning framework calls for an efficient planner that is inherently aware of both user preferences and real-world feasibility constraints.


To address these issues, this study develops a framework that integrates LLMs and preference-aware heuristic algorithms for MUTIP. 
Specifically, we design a dual-step LLM-based preference analysis pipeline that translates a user's natural-language request into structured algorithm inputs and personalized preference signals. These signals are then injected into a GRASP-based planning procedure~\parencite{exposito2019fuzzy}, enhanced with a ``Clustering and Substitution'' strategy to improve route rationality and efficiency under multi-day constraints. This integration yields itineraries that are both controllable (constraint-feasible) and aligned with user demand.

In summary, the key contributions of this paper are as follows:
\begin{itemize}
    \item We have devised an end-to-end framework that seamlessly integrates LLM-based preference extraction with preference-aware itinerary planning algorithm to address MUTIP, enabling controllable multi-day itineraries from a single natural-language interaction.
    \item We develop a dual-step LLM-based preference analysis pipeline and enhance a GRASP-based planner with a ``Clustering and Substitution'' strategy to improve itinerary alignment, rationality, and efficiency under multi-day constraints.
    \item We proposed metrics to evaluate the quality and rationality of travel itineraries, and validated our framework using two real-world datasets. Our framework significantly outperforms representative LLM-based methods, improving the average total itinerary score by at least 4.52\% and 11.09\% across 5,040 user cases with diverse preferences in the two datasets. Furthermore, through end-to-end algorithmic enhancements, it achieves notable average improvements of 17.95\% and 26.07\% in the computed metrics, while also delivering substantial gains in time efficiency, realizing average performance increases of 4.64\% and 25.55\% within shorter computation times compared to suboptimal methods that require multiple iterations. These results demonstrate the superior quality, controllability, and computational efficiency of our approach.
\end{itemize}

\section{Related Work}
Given the relevance to the study presented in this work and the evolution of technology, we delve into a discussion on both traditional itinerary planning techniques and itinerary planning based on LLMs, respectively.
\subsection{Traditional Travel Itinerary Planning} 
Traditional research~\parencite{urrutia2021variable,Hu_Fathi_Pardalos_2018} on travel itinerary planning often leverages concepts such as the OP and the TOP, employing various heuristic algorithms~\parencite{lin2017solving, rezki2018lambda} to tackle different variants of these problems. For instance, ~\textcite{exposito2019fuzzy} utilized the GRASP algorithm to optimize the overall scores of personalized itineraries, taking into consideration the diverse categories of tourist attractions. Similarly, ~\textcite{tlili2021simulated} developed the KSA algorithm, which integrates k-means clustering with simulated annealing~\parencite{bertsimas1993simulated} to generate multi-day travel itineraries, incorporating it into a Travel Recommendation System (TRS) named StayPlan. \textcite{derya2024selective} addressed itinerary planning involving clustered attractions and demonstrated the effectiveness of their algorithms for this specific variant of the itinerary planning problem.

As research progresses, some studies have broadened the scope of optimization by incorporating multi-objective optimization techniques into the problem~\parencite{ruiz2021multi,piya2023optimization}. For example, ~\textcite{aliano2024effective} focused on minimizing travel distance while maximizing service levels through multi-objective optimization utilizing the Chebyshev Distance. \textcite{kolaee2024sustainable} and~\textcite{pitakaso2024multi} introduced sustainability considerations into their models, incorporating environmental and social factors. 

Despite these advancements, much of the existing research prioritizes algorithmic efficiency and optimality, frequently overlooking the intuitive user experience and practical usability of the developed systems. To enhance the user experience, several studies have sought to incorporate user preferences into their models. \textcite{DBLP:journals/ipm/BrilhanteMNPR15} treated photos as traces of tourist behavior and extracted spatio-temporal information from them to analyze sightseeing itineraries. \textcite{paulavivcius2023novel} incorporated hierarchical constraints and personalized ratings of POIs into the TTDP, developing a TRS that enables users to set preferences through predefined options and calculates personalized scores for POIs based on their characteristics. \textcite{zhou2023tour} integrated user interest-driven POI recommendations into itinerary planning by combining traditional machine learning algorithms, such as Naive Bayes and decision forests. \textcite{DBLP:journals/ipm/OrabiAA25} integrated both static and dynamic sources of POI lists to address transient and upcoming attractions.

However, these algorithms rely significantly on predefined user options or intrinsic attributes of POIs, which constrains their adaptability to diverse user needs and complicates their application in complex settings. 

\subsection{LLMs-based Travel Itinerary Planning} 

Due to the limitations inherent in traditional itinerary planning algorithms, several studies have explored integrating LLMs, which are proficient in language comprehension and generation, into this task. \textcite{xie2024travelplanner} introduced a benchmark dataset named TravelPlanner, which integrates flight data, driving distances and times, restaurant information, tourist attractions, and accommodation details. This dataset is designed to test the practical application capabilities of language models in addressing cross-city itinerary planning problems. For this benchmark, ~\textcite{hao2024large} proposed a framework that formulates the travel planning problem as a Satisfiability Modulo Theories (SMT) problem, solved by integrating LLMs with SMT solver, effectively leveraging the natural language parsing capabilities of LLMs and the multi-constraint problem-solving abilities of SMT solvers. Similarly, ~\textcite{gundawar2024robust} introduced the LLM-Modulo framework, combining LLMs with external validators through an iterative process to improve LLM performance on the TravelPlanner task.
Moreover,~\textcite{ju2024globe} employed a fine-tuned Llama model to convert natural language inputs into JSON format, followed by solving the planning problem through mixed-integer linear programming. \textcite{shao-etal-2025-personal} introduced RealTravel, an augmented version of the TravelPlanner, and proposed an integrated system that combines LLMs with numerical solvers to generate travel plans.

Unlike the cross-city itinerary planning problem, for single-city planning, \textcite{de2024trip} employed LLMs to acquire and translate travel information, then 
utilized automated planners to generate travel plans. These plans ensure constraint satisfaction and optimize objectives by combining the strengths of both technologies to produce effective single-day travel itineraries. \textcite{tang2024synergizing} utilized a combination of LLMs and traditional spatial optimization models to address the open-domain urban itinerary planning (OUIP) task. This task facilitates single-day, city-walk type itinerary planning based on  natural language descriptions of users. 

\newcolumntype{Y}{>{\centering\arraybackslash}X}
\begin{table*}[h]
\centering
\caption{Systematic comparison of the proposed methods and baselines. The comparison is presented in two vertically aligned sections to maintain readability while covering all critical requirements.}
\label{tab:comparisonr}
\footnotesize

\begin{minipage}{\textwidth}
    \begin{tabularx}{\textwidth}{p{2.6cm} | YYYYY | YYYY}
    \toprule
    \multirow{2}{*}{\textbf{Method}} & \multicolumn{5}{c|}{\textbf{Base Route Constraints}} & \multicolumn{4}{c}{\textbf{Personalization}} \\
    \cmidrule(lr){2-6} \cmidrule(lr){7-10}
    & Dur. & Win. & Man. & Mul. & Bud. & Num. & Obj. & Prof. & Scor. \\
    \midrule
    GlobeTrott~(\citeyear{paulavivcius2023novel}) & \checkmark & --- & \checkmark & --- & --- & --- & --- & \checkmark & \checkmark \\
    StayPlan~(\citeyear{tlili2021simulated}) & \checkmark & --- & --- & \checkmark & --- & --- & --- & --- & --- \\
    FuzzyGRASP~(\citeyear{exposito2019fuzzy}) & \checkmark & \checkmark & \checkmark & \checkmark & --- & \checkmark & \checkmark & --- & --- \\
    TravelPlanner~(\citeyear{hao2024large})& \checkmark & --- & --- & \checkmark & \checkmark & --- & --- & --- & --- \\
    ITINERA~(\citeyear{tang2024synergizing}) & \checkmark & --- & \checkmark & --- & --- & \checkmark & --- & --- & \checkmark \\
    PTS~(\citeyear{shao-etal-2025-personal}) & \checkmark & --- & --- & \checkmark & \checkmark & \checkmark & \checkmark & --- & \checkmark \\
    DeepTravel~(\citeyear{ning2025arxiv}) & --- & --- & \checkmark & \checkmark & \checkmark & --- & --- & --- & --- \\
    \midrule
    \textbf{UDOIP (Ours)} & \checkmark & \checkmark & \checkmark & \checkmark & \checkmark & \checkmark & \checkmark & \checkmark & \checkmark \\
    \bottomrule
    \end{tabularx}


    \begin{tabularx}{\textwidth}{p{2.6cm} | YYYYYY | p{1.4cm} | p{1.8cm} | p{0.7cm}}
    \multirow{2}{*}{\textbf{Method}} & \multicolumn{6}{c|}{\textbf{Logistics \& Logic}} & \multirow{2}{*}{\makecell[l]{\textbf{Data Util.}}} & \multirow{2}{*}{\textbf{Algorithm}} & \multirow{2}{*}{\textbf{Scope}} \\
    \cmidrule(lr){2-7}
    & Cls. & P.S. & In- & Ex- & Din. & Hot. & & & \\
    \midrule
    GlobeTrott (\citeyear{paulavivcius2023novel}) & \checkmark & --- & \checkmark & --- & --- & --- & Quest. & Genetic & Single\\
    StayPlan (\citeyear{tlili2021simulated}) & \checkmark & \checkmark & \checkmark & --- & \checkmark & \checkmark & UI Select & K-means+SA & Single\\
    FuzzyGRASP (\citeyear{exposito2019fuzzy}) & \checkmark & \checkmark & \checkmark & --- & \checkmark & \checkmark & Predefined & Fuzzy GRASP & Single\\
    TravelPlanner (\citeyear{hao2024large}) & \checkmark & --- & \checkmark & \checkmark & \checkmark & \checkmark & Query Ext. & LLM+SMT & Cross\\
    ITINERA (\citeyear{tang2024synergizing}) & \checkmark & --- & --- & --- & --- & --- & Query Ana. & LLM+CSO & Single\\
    PTS (\citeyear{shao-etal-2025-personal}) & \checkmark & \checkmark & --- & \checkmark & \checkmark & \checkmark & Reviews & LLM+SASRec & Cross\\
    DeepTravel (\citeyear{ning2025arxiv}) & --- & --- & \checkmark & \checkmark & --- & \checkmark & Query Ext. & Agentic RL & Cross\\
    \midrule
    \textbf{UDOIP (Ours)} & \checkmark & \checkmark & \checkmark & --- & \checkmark & \checkmark & \textbf{Implicit} & LLM+GRASP & Single\\
    \bottomrule
    \end{tabularx}
\end{minipage}

\begin{flushleft}
\scriptsize
\textbf{Note:} \textbf{Dur.}: Tour Duration; \textbf{Win.}: Time Window; \textbf{Man.}: Mandatory POIs; \textbf{Mul.}: Multi-day; \textbf{Bud.}: Budget; \textbf{Num.}: POI Count; \textbf{Obj.}: Multi-objective; \textbf{Prof.}: User Profile (e.g. slow pace); \textbf{Scor.}: Profile-dependent Scores; \textbf{Cls.}: POI Classification; \textbf{P.S.}: POI Scores; \textbf{In-/Ex-}: Intra/Inter-city Transport; \textbf{Din.}: Dining; \textbf{Hot.}: Hotel; \textbf{Data Util.}: Data Utilization Methods; \textbf{Algorithm}: Solution Algorithm; \textbf{Scope}: Single-city or Cross-city(a one-way flight from origin city A to destination city B, with tourism activities confined exclusively to city B).
\end{flushleft}
\end{table*}

While the aforementioned studies have made significant progress in leveraging LLMs for both cross-city and single-city itinerary planning, they primarily focus on single-day or constraint-specific tasks, leaving a gap in addressing the more complex challenges of multi-day urban travel itinerary planning which requires a nuanced understanding of evolving user preferences. Only by bridging this gap can the resulting itineraries achieve practical usability and be genuinely adopted by users in real-world travel scenarios.To more clearly illustrate the distinctions between our work and others, as well as highlight our advantages, we have conducted a systematic comparison. As shown in Table \ref{tab:comparisonr}, our approach explicitly considers and addresses multiple critical requirements of real-world travel that prior methods have failed to handle -- this is also reflected in the experimental section.

In this work,
we concentrate on the diverse demands of users and develop a more flexible framework for the MUTIP problem. Our proposed method allows users to quickly obtain a personalized, high-quality itinerary that is both humanized and highly specific through a convenient natural language description. This method enhances the controllability and usability of existing approaches, more accurately aligning with user preferences and system constraints.

\section{Preliminary}
In this section, we define the Multi-day Urban Travel Itinerary Planning (MUTIP) problem and introduce the GRASP algorithm along with the LLM technique that underpins our framework.

\subsection{Multi-day Urban Travel Itinerary Planning (Research Objective)}

Given that urban travel often involves numerous complex POIs and should extend beyond a single day, we define the MUTIP problem. The objective is to select prioritized POIs and generate multiple routes over several days based on the user's demand description. Users express their travel planning demands \( S_u \) through natural language, which may include specifying a target destination city \( C_u \), requirements for POIs \( P_u \), time constraints \( T_u \), dining preferences \( D_u \),  and other considerations. To accommodate these demands, a POI database for each city is constructed by integrating information from tourism platforms and relevant map service APIs, such as TripAdvisor\footnote{https://www.tripadvisor.com/} and Google Maps\footnote{https://developers.google.cn/maps}. Each POI \( p \) contains essential data, including its name \( n_p \), type \( c_p \), description \( d_p \), geographical coordinates (latitude \( g^{lat}_p \) and longitude \( g^{long}_p \)), suggested visiting duration \( t^{visit}_p \), operating time window \( t^{window_{open}}_p \), rating \( r_p \), and the number of ratings \( n^r_p \). The travel time between two POIs \( t^{traffic}_{(p_1, p_2)} \) is also obtained through map services. The notations used in this paper, along with their corresponding descriptions, are summarized in Table~\ref{tab:symbols} for clarity. Note that the user requirements we have listed are not mandatory; instead, the system default settings can be utilized in their absence. However, this paper focuses on addressing the critical issue of how to better accommodate users' personalized needs once they are presented.

\begin{table}[ht]
\centering
\caption{Summary of Notations}
\begin{tabular}{c|l}
\hline
\textbf{Symbol}          & \textbf{Description} \\ \hline
$S_u$                    & User's travel planning demands \\ 
$C_u$                    & User-specified target destination city \\ 
$P_u$                    & User's requirements for Points of Interest (POIs) \\ 
$T_u$                    & User's time constraints for the itinerary \\ 
$D_u$                    & User's dining preferences during the itinerary \\ 
$k$                    & Total number of days of the itinerary specified by the user \\ 
$p$                      & A point of interest (POI) \\ 
$n_p$                    & Name of the POI $p$ \\ 
$c_p$                    & Category of the POI $p$ \\ 
$d_p$                    & Description of the POI $p$ \\ 
$g^{lat}_p, g^{long}_p$  & Geographical coordinates (latitude and longitude) of the POI $p$ \\ 
$t^{visit}_p$            & Suggested visiting duration for the POI $p$ \\ 
$t^{window_{open}}_p$    & Operating time window of the POI $p$ \\ 
$r_p$                    & Rating of the POI $p$ \\ 
$n^r_p$                  & Number of ratings for the POI $p$ \\ 
$t^{traffic}_{(p_1, p_2)}$ & Travel time between two POIs $p_1$ and $p_2$ \\ 
$L_{P_u}$                & Set of required (prioritized) POIs for the user \\ 
$t^{start}_u$            & User's desired or specified start time each day \\ 
$t^{plan}_u$             & Maximum planned travel time per day \\ 
$t^{visit}_{(p, u)}$            & Recommended visiting time for each POI for a particular user $u$\\ 
$n^{c}_u$                & Total number of dining-related POIs per day \\ 
$s_{(p,u)}$                    & Score of the POI $p$ for a particular user $u$ used for itinerary optimization\\ 
$w_u$                    & Visit duration compactness parameter for a particular user $u$ \\ 
$RCL$                    & Restricted Candidate List used in GRASP for initial solution construction \\ 
$RCL^{*}$                & Refined Restricted Candidate List after applying fuzzy rule \\ \hline
\end{tabular}
\label{tab:symbols}
\end{table}

Our goal is to develop a method that generates a specific multi-day travel itinerary, including a complete sequence of POIs for each day, with suggested arrival time and visiting duration. 
This itinerary must adhere to system constraints, such as POI Opening hours, while also satisfying user demands \( S_u \), such as preferences for specific types of attractions, personalized time management, and dietary needs, which are further classified into POI demands \( P_u \), time constraints \( T_u \), and dining demands \( D_u \). For POI demands \( P_u \), we consider the required POIs \( L_{P_u} = \{p^1_u, p^2_u, ..., p^l_u\} \) and the score of the $n$ POIs in the database $\{s_{(p_1, u)}, s_{(p_2,u)}, ..., s_{(p_n, u)}\}$ computed from $r_p$, $n^r_p$ and $L_{P_u}$, which is detailed in Section \ref{method:icp} and formalized in Formula (\ref{eq:score}).
For time constraints \( T_u \), we focus on the total number of days \( k \) specified, the user's desired or specified start time each day \( t^{start}_u \), the maximum planned travel time per day \( t^{plan}_u \), and the recommended visiting time \( t^{visit}_{(p,u)} \) for each POI $p$ and a particular user $u$. For dining demands \( D_u \), we consider the total number of dining-related POIs per day \( n^{c}_u \). 

\subsection{Greedy Randomized Adaptive Search Procedure}
\label{preliminary-grasp}

The Greedy Randomized Adaptive Search Procedure (GRASP)~\parencite{marques1999grasp} is an iterative heuristic algorithm that combines greedy strategies with randomization techniques to find near-optimal solutions. The core concept of GRASP involves constructing an initial solution during each iteration, followed by improvement through local search. The algorithm consists of two main steps:
\begin{enumerate}[(1)]
\item \textbf{Solution Construction:} In each iteration, the algorithm greedily selects an element from the candidate set to add to the current solution. This selection process employs a Restricted Candidate List (RCL) strategy, which considers only the top subset of candidates. The selection is repeated until a complete solution is achieved. In itinerary planning algorithms, a greedy strategy is typically employed to prioritize the insertion of POIs that result in the shortest increase in overall travel time.
\item \textbf{Local Search:} After constructing the initial solution, local search is employed to refine it, typically involving neighborhood operations such as swaps.
\end{enumerate}

Traditional GRASP typically optimizes a single metric, such as minimizing time. Recognizing the consistency between minimizing time and maximizing the total POI score objective, \textcite{exposito2019fuzzy} introduced the Fuzzy-GRASP algorithm for itinerary planning. This algorithm employs a greedy time function to determine optimal solutions for each POI during the construction of RCL. Before randomly selecting an element from the RCL, the list is sorted by the POI scores \( s_p \) and truncated to create a refined list RCL*. This fuzzy strategy prioritizes POIs with higher scores. 

Overall, the Fuzzy-GRASP process begins by generating an initial solution using a greedy heuristic strategy, followed by a local search to further optimize the solution. In each iteration, a RCL is constructed, and a fuzzy rule is applied to filter it, yielding a refined list RCL*. From this refined list, a solution is randomly selected and subsequently optimized through neighborhood search techniques. By explicitly accounting for multiple essential real-world constraints inherent to itinerary planning (such as time windows) -- this approach establishes a robust foundation that motivates our work. Our work draws inspiration from this approach and introduces enhancements aimed at improving itinerary optimization, increasing adaptability to diverse user preferences, and improving the efficiency of the algorithm.

Specifically, in our enhanced based planning algorithm, we executed the Solution Construction and Local Search phases in an alternating manner
several times: each time the solution construction is completed, the algorithm immediately proceeds to the local search phase to optimize the current solution. During local search, a more time-efficient sequence of solutions is identified, allowing the subsequent construction phase to build upon the existing solution rather than restarting from scratch. This iterative process continues until a stopping criterion, such as a predefined number of iterations or the attainment of a specific solution quality, is satisfied, thereby progressively converging toward a globally optimal solution.


\section{Methodology}
\label{sec:method}
\begin{figure*}[h]\centering
    \includegraphics[width=1.0\columnwidth]{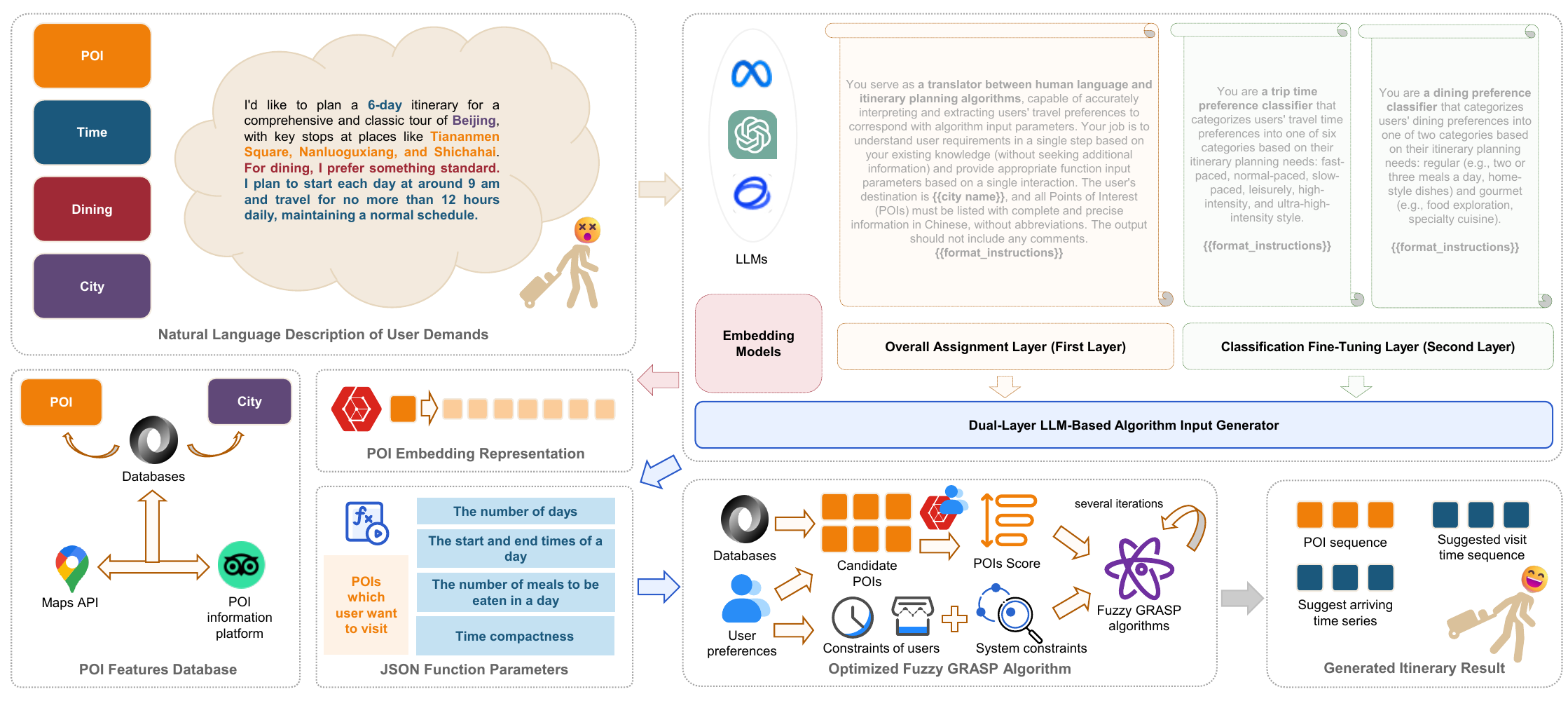}
	\caption{Framework of User Demand Oriented Itinerary Planner (UDOIP)}
    \label{FIG_1}
\end{figure*}

Fig.~\ref{FIG_1} illustrates the overall framework of our proposed method, named the User Demand-Oriented Itinerary Planner (UDOIP). This framework primarily comprises four key components: user requirements gathering, multi-source data integration, solution architecture design, and final results delivery.
To solve the MUTIP problem, our framework captures user preferences across three basic categories: POIs, time, and dining, and employs a dual-step LLM-based process for understanding personalized preferences. This process begins with LLMs interpreting user demands, followed by the application of a GRASP-based heuristic planner enhanced with constraint- and preference-aware strategies to 
generate practical and optimized multi-day travel plans that satisfy user-specific constraints within these categories. 

In this section, we introduce the two crucial modules of the framework: the LLM-based algorithm input generator and the GRASP-based planner 
 for itinerary planning. Additionally, in the final subsection, we summarize all the methodological components and present the overall process of our framework.

\subsection{LLMs-based algorithms inputs generator}
\label{method-llms}

Before executing the planning algorithm, personalized input settings for each user are necessary. However, users often find complex function input configurations difficult to understand and tedious to set up individually. They prefer to describe their requirements directly rather than adjust parameters one by one. Therefore, implementing a system that supports natural language input is crucial. One approach is to guide the LLM by defining the meaning and configuration rules for each input parameter, enabling it to generate algorithm inputs directly and integrate seamlessly with the system.

However, when dealing with numerous input parameters, the LLM may struggle to accurately interpret specific parameters, even occasionally leading to misunderstandings. To address this challenge, we propose a dual-layer LLM framework, as shown in the upper right part of Fig.~\ref{FIG_1}. The first layer, the overall assignment layer, generates values for all input parameters based on the user's descriptions, covering both enumerated parameters (e.g., prioritized lists of POI) and fixed parameters (e.g., visit duration compactness parameter). The second layer, known as the classification fine-tuning layer, provides macro-level oversight of the outputs from the first layer, categorizing user preferences in areas such as time and dining, and adjusting values that significantly deviate from expected ranges. This dual-layer approach preserves the fine-grained understanding of user preferences while reducing inaccuracies in parameter comprehension, ultimately enhancing the model's ability to accurately capture and respond to user demands.

To furnish the itinerary planning algorithm with necessary input parameters, which will be elaborated on in Section \ref{opt-grasp}, we leveraged the capabilities of LLM to identify requisite input parameters from the user's natural language description. 
The detailed descriptions and types of the $7$ main parameters relevant to user demands are summarized in Table~\ref{table:parameters}.


\begin{table}[h]
\centering
\caption{Main algorithm input parameters}
\label{table:parameters}
\begin{tabularx}{\textwidth}{l|c|X|c}
\toprule
\textbf{Preference} & \textbf{Parameter} & \textbf{Description} & \textbf{Type} \\
\midrule
POI & \( L_{P_u} \) & Set of required (prioritized) POIs for the user & \texttt{enum} \\
\midrule
\multirow{4}{*}{time}
    & \( k \) & Total number of days of the itinerary specified by the user & integer \\
    & \( w_u \) & Visit duration compactness parameter for a particular user $u$ & decimal \\
    & \( t^{\mathrm{start}}_u \) & User's desired or specified start time each day & datetime \\
    & \( t^{\mathrm{plan}}_u \) & Maximum planned travel time per day (in hours) & decimal \\
\midrule
    \multirow{2}{*}{dining} 
    & \(\text{min } n^{c}_u \) & Minimum number of dining-related POIs per day & integer \\
    & \(\text{max }  n^{c}_u \) & Maximum number of dining-related POIs per day & integer \\
\bottomrule
\end{tabularx}
\end{table}

Taking POI preferences as an example, the corresponding parameter is \( L_{P_u} \), which could be empty as well. Since this parameter is of an enumerated type, it can be fully processed by the first layer of the LLM, without the need for further fine-tuning. 

For time preferences, the relevant parameters include \( t^{start}_u \), \( t^{plan}_u \), and \( w_u \), whose usage is detailed in the next subsection. These parameters can be directly assigned values by the first layer of the LLM according to each user query. In the second layer, user demands are classified into categories such as ``fast-paced'', ``normal-paced'', ``slow-paced'', ``leisure time'', ``high-intensity time'' and ``ultra-high-intensity style''. For each category, normal ranges for the related parameters are predefined manually. If any parameter values generated by the first layer fall outside the typical range for the assigned category in the second layer, they are adjusted to align closer to the standard range. For dining-related user preferences, a similar approach is applied. When users express a preference for exploring gourmet experiences, LLMs assist in adjusting \(\text{min } n^{c}_u \) and \(\text{max } n^{c}_u \). Otherwise, the subsequent process follows the standard choices preferred by the majority of users. Detailed prompts are provided in Appendix~\ref{ap-prompts}, and the complete process is outlined in Algorithm~\ref{alg:llm-input-gen} in Appendix \ref{ap:alg}. As demonstrated in the experiments detailed in Section~\ref{exp-comparison}, integrating the second-layer LLM classifier significantly enhances the LLM's capacity to accurately capture user preferences.


 \subsection{GRASP-based Planning Algorithms}
\label{opt-grasp}

\begin{figure*}[h]\centering
        \includegraphics[width=0.9\columnwidth]{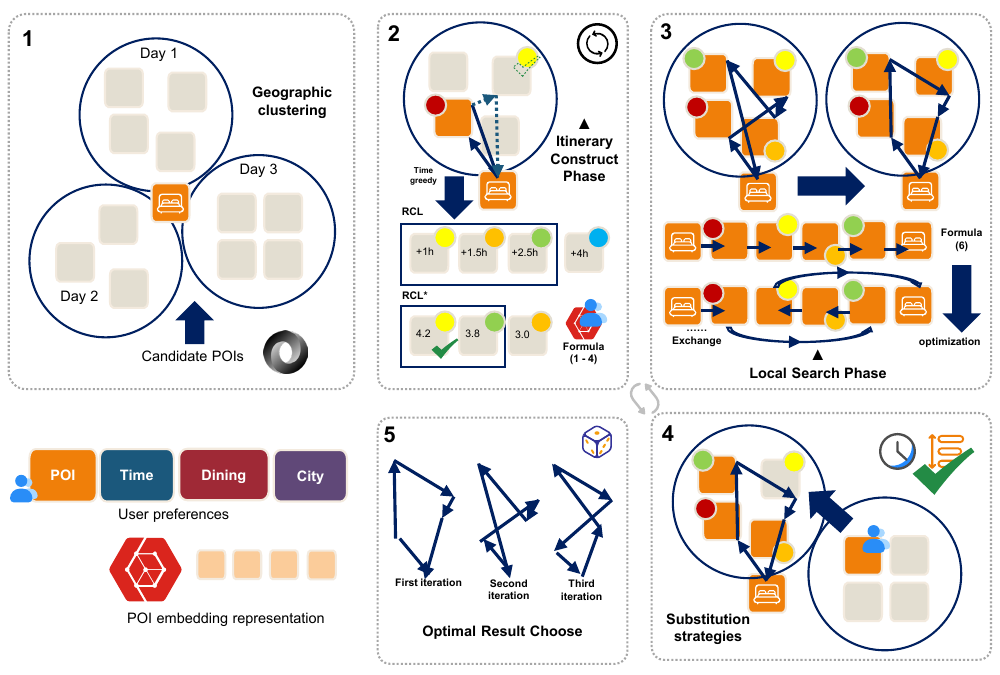}
	\caption{Framework of GRASP-based Planning Algorithms}
    \label{fig:method-grasp}
\end{figure*}

Through the precise understanding of user preferences enabled by the aforementioned LLMs, we obtained various variables that effectively represent user demands. 
However, directly outputting these results as itinerary planning outcomes poses several critical issues. The generated itineraries may violate system constraints, require substantial post-processing before they can be practically used by travelers, or even contain hallucinated entities that do not exist in the underlying database. This last issue is empirically demonstrated in Section \ref{exp-comparison}.
Therefore, in this section, we propose an enhanced approach to fulfilling the personalized needs of each user by extending the fundamental principles of the GRASP algorithm introduced in Section \ref{preliminary-grasp}. By incorporating user constraints into the framework, we have developed several extensions of GRASP that operates collaboratively to address diverse individual requirements. The detailed interplay and operation of the two main phases, along with the tailored enhancement strategies supplementing them, are meticulously described as follows to illustrate how they work together to ensure that user-specific constraints are seamlessly integrated into the design and implementation process. 

\subsubsection{Itinerary Construct Phase:}
\label{method:icp}
As shown in Fig.~\ref{fig:method-grasp}, in the MUTIP problem, the initial itinerary typically consists of $k$ arrays, each representing the itinerary for a particular day. These arrays incorporate either a central hotel location or a user-specified hotel as the starting and ending points. According to the GRASP algorithm, each iteration evaluates potential POIs using a greedy time-based approach to identify the optimal insertion point within the current itinerary, i.e., a position where inserting the POI causes the smallest increase in travel duration compared to other feasible positions. Each POI, paired with its best insertion position, is added to the RCL. Subsequently, the RCL is sorted based on the increasing time associated with each tuple, retaining only the $RCL\_SIZE$ tuples. A tuple is then randomly selected from this sorted list, and the POI, along with its optimal insertion position, is inserted into the current itinerary. 
Following each insertion, previous position calculations may no longer be accurate, necessitating a reconstruction of the RCL and a continuation of the random selection process to fill the itinerary. This process repeats until no further POIs can be included—either because the maximum allowable playtime is reached or all POIs have been integrated.
It is evident that tuples within the RCL have a high likelihood of being incorporated into the itinerary. Conversely, tuples with low user preference priorities, or those failing to meet user or system constraints, should be excluded from the RCL. Thus, constructing an effective RCL is essential to the algorithm's success.

In the Fuzzy GRASP algorithm, a key enhancement lies in constructing the RCL* by incorporating POI scores as an additional consideration, rather than relying solely on the minimal increase in travel time. One method involves directly employing average ratings sourced from online data as POI scores. However, 
this method may overlook popularity metrics and lack statistical significance when the number of ratings is low or limited. In addition, conventional scoring methods also fail to consider user preferences, resulting in suboptimal selections of POIs that do not align with user interests. Therefore, to comprehensively evaluate POI scores while incorporating user preferences, we propose the Personalized LLM-enhanced POI Scoring (PE-PS) method, which employs the following formulas:

\begin{equation} \label{eq:emb_poi}
emb_p = LLM^{embedding}(n_p + d_p)
\end{equation}
\begin{equation} \label{eq:average_emb}
emb_u = \text{Average}([emb_{p^1_u}, emb_{p^2_u}, \ldots, emb_{p^l_u}])
\end{equation}
\begin{equation} \label{eq:similarity}
sim_{(p,u)} = \frac{emb_p \cdot emb_u}{\|emb_p\| \cdot \|emb_u\|}
\end{equation}
\begin{equation} \label{eq:score}
s_{(p,u)} = r_p \cdot \log_2(n^r_p + 1) \cdot sim_{(p,u)}
\end{equation}

In this context, Formula (\ref{eq:emb_poi}) employs Large Language Models to embed pertinent POI information, encompassing the POI name and description, into a high-dimensional representation.
Next, Formula (\ref{eq:average_emb}) computes each user's representation 
by averaging the representations 
of 
POIs preferred by the user. Formula (\ref{eq:similarity}) calculates the cosine similarity between each user-POI pair, utilizing their corresponding representations. 
Finally, Formula (\ref{eq:score}) integrates 
the rating, the number of ratings, and the similarity score to ascertain the comprehensive 
score of each POI for any given 
user. This method allows the framework to more accurately rank POIs by not only considering their popularity and ratings but also aligning with user preferences, thus solving the problem of selecting POIs that better match the user's interests while accounting for varying levels of statistical significance in the ratings data. In this scenario, a specific POI can exhibit diverse scores to various users.
In our experiments in Section~\ref{sec-exp-case}, we validated the effectiveness of PE-PS.

However, unconditionally placing all (POI, optimal insertion position) pairs with short additional travel time and high scores into the RCL* presents certain issues. For instance, prioritizing the insertion of sightseeing POIs may occupy most of the available restaurant hours, thereby failing to meet the user's dining demands. Similarly, inserting POIs not explicitly mentioned by the user early in the process may result in insufficient time for more preferred POIs. Therefore, it is necessary to perform an initial screening of candidate POIs before constructing the RCL*, based on the following conditions:

\begin{enumerate}
    \item Priority should be given to satisfying the user's minimum dining needs min \( n^{c}_u \) and preferred POIs \( L_{P_u} \). When these two categories of demands are not satisfied at the initial stage, other unrelated POIs should not be inserted.
   
    \item Evaluate whether the insertion of a POI exceeds the maximum daily travel time \( t^{plan}_u \). If it does not, assess whether the insertion fits within an available time window. If multiple windows are available, select the one with the shortest waiting time.
    
    \item When the RCL* becomes empty, indicating that no additional POIs can be inserted into the current itinerary, the local search phase is immediately executed to assess whether the itinerary duration can be further minimized. This is followed by a new round of RCL* construction to verify if the RCL* remains empty.
    
    \item If the RCL* is empty for two consecutive alternating loops, it is determined that user demands cannot be fully met. At this point, priority constraints are relaxed, allowing all POIs to be considered for insertion rather than only those in the initial priority list. Naturally, care must be taken to ensure that the number of restaurant-type POIs does not exceed the user's expected maximum dining limit \( \text{max } n_u^c \), along with other similar maximum-type constraints.
\end{enumerate}

Additionally, to better accommodate various real-world scheduling demands, we introduced a visit duration compactness parameter, \( w_u \), which correlates with the user's preferences and is initialized by LLMs and dynamically adjusted throughout the GRASP process. When the total available time is evidently insufficient to satisfy the user's desire to visit a larger number of POIs, this parameter is gradually reduced within a certain range to enable the user to visit more of their desired POIs. The relationship governing this adjustment is defined by Formula (\ref{eq:timevisituser}):

\begin{equation} \label{eq:timevisituser}
t^{visit}_{(p,u)} = w_u \cdot t^{visit}_p
\end{equation}

We describe the detailed steps of this phase in Algorithm~\ref{alg:rcl-construct} in Appendix \ref{ap:alg}. These steps ensure that the POI selection process optimizes the overall travel experience by aligning with user preferences while ensuring practical constraints, such as maximum travel time of one day and open time windows of POIs. Consequently, the algorithm is capable of generating itineraries that are both feasible and aligned with the user's priorities, minimizing the risk of time conflicts or unmet demands.

\subsubsection{Local Search Phase:} During each iteration of the local search phase, the algorithm seeks to identify a locally optimal solution for the given itinerary plan. General algorithms predominantly relied on a single greedy time-based strategy, neglecting the impact of the itinerary on the user's experience during the actual travel process. Hence, to quantify the quality of a generated itinerary and improve user satisfaction, we introduce an optimization objective function that incorporates three critical factors: waiting time, optimal time window, and travel time. The optimization objective function is expressed as follows:

%
%

\begin{equation} \label{eq:evaluation}
F_{\text{local}} = \left( 1 + \omega_{\text{std}} + N_{\text{mis}} / N_{\text{opt}} \right) \cdot \left( \beta \cdot W_{\text{total}} + T_{\text{total}} \right), \beta \in [0, 1]
\end{equation}

Here, \( \omega_{\text{std}} \) denotes the normalized standard deviation of waiting times, used to penalize variations in waiting time across POIs.  \( W_{\text{total}} \) is the total waiting time, defined as the time the user must wait for a POI to open. This typically occurs when the user arrives before the POI's opening hours. Reducing waiting time or averaging it are keys to minimizing idle periods in the itinerary. The travel time \( T_{\text{total}} \) refers to the total time spent traveling between POIs. Minimizing travel time is essential to improving the overall efficiency of the itinerary, as it reduces unnecessary commuting and allows users to spend more time at destinations. In this function, \( \beta \) is a hyperparameter that adjusts the relative importance of waiting time \( W_{\text{total}} \) in comparison to travel time \( T_{\text{total}} \), with values ranging between 0 and 1. 

The term \( N_{\text{mis}} \) represents the number of POIs where the visit does not align with the optimal time window, while \( N_{\text{opt}} \) denotes the total number of POIs that can be visited during their optimal windows. The optimal time window refers to the ideal time to visit a POI based on contextual factors, such as specific viewing times (e.g., sunrise/sunset point of time) or mealtimes for dining-related POIs. Aligning visits with these optimal windows ensures a higher-quality user experience by enhancing the enjoyment of POIs. 


The overall objective of the optimization function is to balance these three factors—waiting time, optimal time window alignment, and travel time—in order to generate a well-optimized itinerary that satisfies user demands while enhancing their actual travel experience. The complete algorithmic process for this phase can be found in Algorithm~\ref{alg:local-search} in Appendix \ref{ap:alg}. It is worth noting that, as introduced in Section~\ref{preliminary-grasp}, this phase alternates a few times with the itinerary construction phase in our algorithm, which differs slightly in its formulation from the traditional GRASP algorithm.

\subsubsection{Clustering and Substitution Strategies}

Parts 2-3 of Fig.~\ref{fig:method-grasp} presents the two phases described above. However, the current algorithm 
tends to generate homogeneous daily itineraries. To address this issue, we introduced a geographic clustering strategy before Parts 2-3 that groups POIs into \( k \) clusters based on their locations, with each cluster assigned to different days. Part 1 of Fig.~\ref{fig:method-grasp} illustrates this step, which ensures greater geographical diversity in the daily itineraries. However, this strategy poses challenges, such as when a large number of prioritized POIs are concentrated within a single cluster. This issue can make it challenging to visit all desired attractions in a single day, particularly if the itinerary strictly adheres to the clustering results. Such adherence could lead to potentially neglecting these attractions on subsequent days, despite the user having reserved additional time.


To mitigate this issue, we introduced priority and substitution strategies. The algorithm initially plans routes within clusters but allows for cross-cluster adjustments if the user's prioritized POIs cannot be included in the itinerary due to clustering constraints.
As shown in Part 
4 of Fig.~\ref{fig:method-grasp}, when no further improvements can be made by adding new POIs, the algorithm attempts to replace non-essential POIs in the current route with unplanned POIs from other clusters. These replacements must involve POIs of the same type, meet system constraints and user constraints, and either be user-specified priority POIs or result in time savings exceeding a threshold
\( \gamma \). By incorporating clustering and substitution strategies, the algorithm can achieve superior results in a shorter time, optimizing both the rationality of routes and the overall score objective. 
Algorithm~\ref{alg:clustering-substitution} in Appendix \ref{ap:alg} illustrates this process. It is important to note that this strategy is not executed sequentially; rather, it is applied across various parts of the entire algorithmic process.

\subsection{Overall Process of UDOIP}

In previous sections, we explained the main components of the UDOIP framework. In this section, we summarize the overall workflow and improvements of our framework using pseudocode. We present the complete workflow of the framework in Algorithm~\ref{alg:main-algorithm-updated}, and the implementation details of specific steps are provided from Algorithms~\ref{alg:llm-input-gen} to~\ref{alg:clustering-substitution}, which are not reiterated here. The only part of the algorithm not previously mentioned is the repeated iteration of the above process to find the optimal solution, as shown in Part 5 of Fig.~\ref{fig:method-grasp}. However, our method focuses on achieving better results with fewer iterations to ensure a seamless user experience in real-time software. Thus, this aspect is not a primary focus of our discussion, which is consistent with the experimental results presented in Section~\ref{exp-ablation}.
We validate this algorithmic pipeline using two real world datasets, with a detailed description of the dataset provided in Section \ref{exp-dataset}.

\begin{breakablealgorithm}
\caption{Algorithm of User Demand Oriented Itinerary Planner}
\label{alg:main-algorithm-updated}
\begin{algorithmic}[1]
\Require User natural language description, POI set  retrieved from the database, time constraints, clustering threshold \( k \), time threshold \( \gamma \), max iterations \( \text{max\_iter} \), fixed iterations \( \text{fixed\_iter} \), RCL\_SIZE, RCL*\_SIZE
\Ensure Optimized itinerary based on user preferences and constraints

\State \textbf{Step: Generate input parameters using LLMs}
\State Call \textproc{LLMs-based-Input Parameters-Generation(user\_description)} (Algorithm \ref{alg:llm-input-gen}) to generate and fine-tune input parameters

\State \textbf{Step: Clustering} 
\State Using \textbf{Kmeans} to group POIs based on geographic location 
\For{$iter \gets 1$ to $\text{max\_iter}$}
    \State Initialize current\_itinerary as empty

    \For{$f\_iter \gets 1$ to $\text{fixed\_iter}$}

        \For {each POI clustering set for each day} 
            \State \textbf{Step: Itinerary Construction}  
            \State Call \textproc{Itinerary-Construct(POI\_clustering\_set, user\_preferences, available\_time, RCL\_SIZE, RCL*\_SIZE, current\_itinerary)} (Algorithm \ref{alg:rcl-construct}) to update current\_itinerary
    
            \State \textbf{Step: Substitution}
            \If{current\_itinerary does not meet all user preferences}
                \State Using \textproc{Substitution Strategy} (Algorithm \ref{alg:clustering-substitution} Lines 3-8) to replace non-essential POIs from other clusters
            \EndIf
        \EndFor

        \State \textbf{Step: Perform Local Search to Improve Solution} 
        \State Call \textproc{Local-Search(current\_itinerary)} (Algorithm \ref{alg:local-search}) to iteratively improve the solution

    \EndFor
    \If{If the $\sum_{k \in K} \sum_{i \in I} s_{(p,u)_i} Y_i^k$ of the current\_itinerary exceeds that of the previously best\_itinerary}
        \Statex \textit{(Here, \( Y_i^k \) is a binary decision variable indicating whether a POI is included in the itinerary. \( K \) represents the set of paths of all days, and \( I \) represents the set of all POIs in every path.)}
        \State Update best\_itinerary by current\_itinerary
    \EndIf
\EndFor


\State Return the final best\_itinerary generated from the above phases

\end{algorithmic}
\end{breakablealgorithm}

\section{Experiment}
\label{sec:exp}

To validate the effectiveness of our proposed approach, we focus on addressing the following research questions:

\textbf{RQ1}: Does our approach provide greater controllability and a deeper understanding of user preferences compared to representative LLM-based baseline?

\textbf{RQ2}: With the addition of a dual-layer LLM architecture in UDOIP, can the method more distinctly differentiate the variations in user preferences compared to a single-layer approach? 

\textbf{RQ3}: Do the additional ``clustering and substitution'' strategy integrated into our method improve the coherence and humanization of the planning results?

\textbf{RQ4}: Can embedding POIs using the LLM's embedding model and calculating POI scores for specific users better align with user preferences?

\textbf{RQ5}: Does our approach demonstrate clear and significant advantages over the recent leading reasoning model, DeepSeek-R1,
when applied to real-world scenarios?

Our experiments are designed in three main stages to address these questions. The experimental setup and results will be detailed in this section.

\subsection{Experimental Setup}
\label{sec-exp-setup}

\subsubsection{Foundation LLM and Datasets}
\label{exp-dataset}


  To control for variability, we use a consistent LLM service (GLM-4-Air\footnote{https://zhipuai.cn/devday}) for all LLM components in UDOIP and for the GLM-based LLM-only baseline. To reflect the current capability frontier of LLM-only planning, we additionally include strong state-of-the-art LLM-only baselines (DeepSeek-V3 and DeepSeek-R1) in RQ1--RQ2. RQ5 further provides a case study with DeepSeek-R1.

 The test datasets are derived from two major Chinese cities, Beijing and Tianjin, both of which are large cities but exhibit significant differences in cultural styles and POI distributions. 
 Additionally, 
 simulated real user demand data constructed based on defined rules by LLMs, which spans all preference categories considered by the algorithm, enabling comparative analysis of itinerary outcomes across different user preferences. The statistical details of the datasets are presented in Table~\ref{table:poi-types} and Table~\ref{table:user-preference}, and specific data samples are provided in Appendix~\ref{ap-datasets} \footnote{Originally presented in Chinese but translated into English.}. 


\begin{table}[h]
\centering
\caption{Numbers of POI Types in Beijing and Tianjin}
\label{table:poi-types}
\begin{tabularx}{\linewidth}{@{}X*{4}{>{\centering\arraybackslash}X}@{}}
\toprule
City & Attr. & Din. & Acc. & Total \\ 
\midrule
Beijing & 104 & 390 & 200 & 694 \\
Tianjin & 68 & 390 & 200 & 658 \\
\bottomrule
\end{tabularx}
\footnotesize{Note: Attr. = Attraction, Din. = Dining, Acc. = Accommodation}
\end{table}

\begin{table*}[h]
\centering
\caption{User Preference Data in Beijing and Tianjin}
\label{table:user-preference}
\begin{tabularx}{\linewidth}{@{}X*{4}{>{\centering\arraybackslash}X}@{}}
\toprule
City & CTT & DT & PT & Total \\ 
\midrule
Beijing & 6 & 2 & 220 & 2640 \\
Tianjin & 6 & 2 & 200 & 2400 \\
\bottomrule
\end{tabularx}
\footnotesize{Note: CTT =The number of covered time preference types , DT = The number of covered dining preference types, PT = The number of covered POI preference types, Total =The number of all user demands data}

\end{table*}

\subsubsection{Baselines and metrics}

To align with the research objectives outlined from RQ1 to RQ4, we selected different baselines and evaluation metrics, which are introduced in detail below.

To answer RQ1, we conducted a comparative study against representative SOTA baselines for MUTIP, selected based on the systematic comparison in Table~\ref{tab:comparisonr}. Specifically, we include (i) LLM baselines that directly generate multi-day itineraries from prompts, and (ii) an adapted solver-assisted baseline, PTS~\parencite{shao-etal-2025-personal}. For the LLM baselines, we keep the prompts consistent with our LLM-based preference extraction and only changing the output format. Detailed prompts are provided in Appendix~\ref{ap-exp-prompts} \footnote{Originally presented in Chinese but translated into English.}. For PTS, since our test dataset and application scenarios lack historical user review data and rely solely on natural language requirements, we adapt the framework to our setting. Specifically, we replace its review-history-dependent re-ranking component with LLM-extracted POI preference keywords matched against our POI database, while keeping the rest of the pipeline unchanged.
Similarly, for RQ2, we use a structure that removes only the second LLM layer as an additional baseline, referred to in the subsequent results as UDOIP (LLM-single-layer).

Three primary evaluation metrics were considered to assess the superior controllability and preference understanding of our method:  

\begin{itemize}
    \item \textbf{Travel Days Satisfaction Failure Rate} (\( R_{\text{days}} \)): This metric represents the proportion of itinerary plans that do not meet the user-specified number of travel days.

    \item \textbf{Average Total Itinerary Score} (\( S_{\text{avg}} \)):
    The total score for a multi-day itinerary is calculated as the sum of the scores of all included Points of Interest (POIs):
    \begin{equation}
    S_{\text{avg}} = \frac{1}{N} \sum_{i=1}^{N} \left( \sum_{j=1}^{M_i} s_{(ij,u)} \right)
    \end{equation}
    where \( N \) is the number of itineraries, \( M_i \) is the number of POIs in the \( i \)-th itinerary, and \(s_{(ij,u)}\) is the score of the \( j \)-th POI in the \( i \)-th itinerary with respect to the user u associated with that itinerary.

    \item \textbf{Preference Variability Test Feature Value(\(H\), \(Z\), and p-value)}:
    We employed the Kruskal-Wallis test~\parencite{mckight2010kruskal} and Dunn's test~\parencite{dinno2015nonparametric} to evaluate the variability in results based on different categories of time and dining preferences, demonstrating the algorithm's capability to capture user preferences. The Kruskal-Wallis test statistic is denoted as \( H \) with p-value \( p\text{-value}_{\text{KW}} \), and Dunn's test statistic is denoted as \( Z \) with p-value \( p\text{-value}_{\text{Dunn}} \). A larger \( H \) and smaller \( p\text{-value}_{\text{Dunn}} \) indicate our method produces more distinct outputs across preference categories than baselines, as quantified in Section \ref{exp-comparison}.
\end{itemize}

In order to enhance the algorithm's ability to generate more rational paths within a shorter time frame, or even within a single iteration, we introduced a ``clustering and substitution'' strategy into the planning algorithm. The effectiveness of these strategies was evaluated using the metrics we developed below to validate RQ3:

\begin{itemize}
    \item \textbf{Total Trip Distance} (\( D_{\text{total}} \)): This metric measures the total length of the planned route, calculated by summing the straight-line distances between consecutive points in the trip:
    \begin{equation}
    D_{\text{total}} = \sum_{i=1}^{n-1} \sqrt{(x_{i+1} - x_i)^2 + (y_{i+1} - y_i)^2}
    \end{equation}
    where \( (x_i, y_i) \) and \( (x_{i+1}, y_{i+1}) \) are the coordinates of consecutive points.

    \item \textbf{Repeated Trip Distance} (\( D_{\text{repeated}} \)): This metric quantifies the length of routes traversed multiple times, often due to loops or overlapping paths:
    \begin{equation}
    D_{\text{repeated}} = \sum \min(d_i, d_{\text{opposite}})
    \end{equation}
    where \( d_i \) is the distance in one direction, and \( d_{\text{opposite}} \) is the distance in the reverse direction.

    \item \textbf{Route Turns} (\( \theta_{\text{turns}} \)): This metric assesses the number and magnitude of turns in the path, reflecting the route's complexity and navigability:
    \begin{equation}
    \theta_{\text{turns}} = \sum \left| \theta_{i+1} - \theta_i \right|
    \end{equation}
    where \( \theta_i \) and \( \theta_{i+1} \) are the angles of consecutive path segments.

    \item \textbf{Multi-day Trip Overlapping Area} (\( A_{\text{overlap}} \)): This metric calculates the overlapping geographical area across multiple trip days, indicating the geographical diversity
    of the routes:
    \begin{equation}
    A_{\text{overlap}} = \sum A_i - A_{\text{union}}
    \end{equation}
    where \( A_i \) is the area of a single trip region, and \( A_{\text{union}} \) is the union area of all trip regions.

    \item \textbf{Time for N Iterations} (\( T_{\text{iter}} \)): This metric represents the total time required (in seconds) for N iterations of itinerary planning, reflecting the algorithm's computational efficiency:
    \begin{equation}
    T_{\text{iter}} = t_N
    \end{equation}
    where \( t_N \) is the total time needed for N iterations.
\end{itemize}

Additionally, to better address RQ3, we compared our complete 
UDOIP framework with several variations below:
\begin{enumerate}[(1)]
    \item LLM-FGRASP, 1 Iteration  
    \item LLM-FGRASP, 2 Iterations  
    \item LLM-FGRASP with Clustering, 1 Iteration  
    \item LLM-FGRASP with Clustering, 2 Iterations  
    \item LLM-FGRASP with Clustering and Substitution, 1 Iteration (UDOIP)  
\end{enumerate}

In the following subsections, we will sequentially present and analyze 
our experimental results.


\subsection{Comparison Experiment (RQ1 RQ2)}
\label{exp-comparison}

\begin{table*}[h] 
\centering
\caption{Performance comparison between our UDOIP and representative SOTA baselines on Beijing and Tianjin datasets}
\label{tab:combined-data}
\resizebox{\textwidth}{!}{
    \begin{tabular}{cccccccc}
    \toprule
    \multicolumn{8}{c}{\textbf{Beijing}} \\
    \midrule
          & $S_{\text{avg}}$ $\uparrow$ & $R_{\text{days}}$ $\downarrow$ & $H_1$ $\uparrow$ & $H_2$ $\uparrow$ & $H_3$ $\uparrow$ & $H_4$ $\uparrow$ & $H_5$ $\uparrow$\\
    \cmidrule{2-8}
    LLM-only (GLM-4-Air) & 83.63 (170.58) & 0.064 & 83.05  & 109.92  & 1627.59  & 69.60  & 110.37  \\
    LLM-only (DeepSeek-V3) & 102.84 (177.63) & 0.042803 & 366.13  & 326.55  & 2411.46  & \textbf{438.98}  & 337.91  \\
    LLM-only (DeepSeek-R1) & 106.98 (201.06) & 0.032197 & 267.93  & 232.91  & 2225.82  & 102.72  & 359.34  \\
    PTS (gpt-4o) & 149.33 (-) & 0     & 0.00  & 15.11  & 0.00  & 0.00  & 0.00  \\
    UDOIP (LLM-single-Layer) & 194.91 (-) & \textbf{0.0015} & 1083.03  & 320.00  & \textbf{2457.60 } & 195.20  & 653.32  \\
    UDOIP (ours) & \textbf{210.15} (-) & \textbf{0.0015} & \textbf{1393.58 } & \textbf{757.95 } & 2419.93  & 401.87  & \textbf{841.86 } \\
    \midrule
    \multicolumn{8}{c}{\textbf{Tianjin}} \\
    \midrule
          & $S_{\text{avg}}$ $\uparrow$ & $R_{\text{days}}$ $\downarrow$ & $H_1$ $\uparrow$ & $H_2$ $\uparrow$ & $H_3$ $\uparrow$ & $H_4$ $\uparrow$ & $H_5$ $\uparrow$\\
    \cmidrule{2-8}
    LLM-only (GLM-4-Air) & 87.12 (150.62) & 0.135 & 60.30  & 69.95  & 1185.83  & 28.61  & 49.93  \\
    LLM-only (DeepSeek-V3) & 76.41 (167.04) & 0.012083 & 138.58  & 150.45  & \textbf{2121.42}  & 79.37  & 25.60  \\
    LLM-only (DeepSeek-R1) & 97.37 (208.47) & 0.027917 & 390.88  & 320.54  & 2083.71  & 74.98  & 77.49  \\
    PTS (gpt-4o) & 129.17 (-) & 0.04125 & 0.00  & 1.79  & 0.00  & 0.00  & 0.46  \\
    UDOIP (LLM-single-Layer) & 208.15 (-) & \textbf{0.0117} & 1345.42  & 333.36  & 2075.67 & 50.25  & 648.27  \\
    UDOIP (ours) & \textbf{231.58} (-) & \textbf{0.0117} & \textbf{1455.93 } & \textbf{949.10 } & 2060.75  & \textbf{596.35 } & \textbf{698.10 } \\
    \bottomrule
    \end{tabular}%
}
\parbox{0.9\linewidth}{\footnotesize{Note: \(H_1\) - H under the average daily usage time, \(H_2\) - H under the average total daily \(t^{visit}_p\), \(H_3\) - H under the average daily \(t^{start}_u\), \(H_4\) - H under \(t^{visit}_p/t^{visit}_{(p,u)}\), \(H_5\) - H under the number of dining type POIs. The numbers in parentheses within $S_{\text{avg}}$ denote the theoretical maximum values.}}
\end{table*}

\begin{figure}[!h] 
    \centering
    \begin{minipage}{0.48\columnwidth} 
        \centering
        \includegraphics[width=\linewidth]{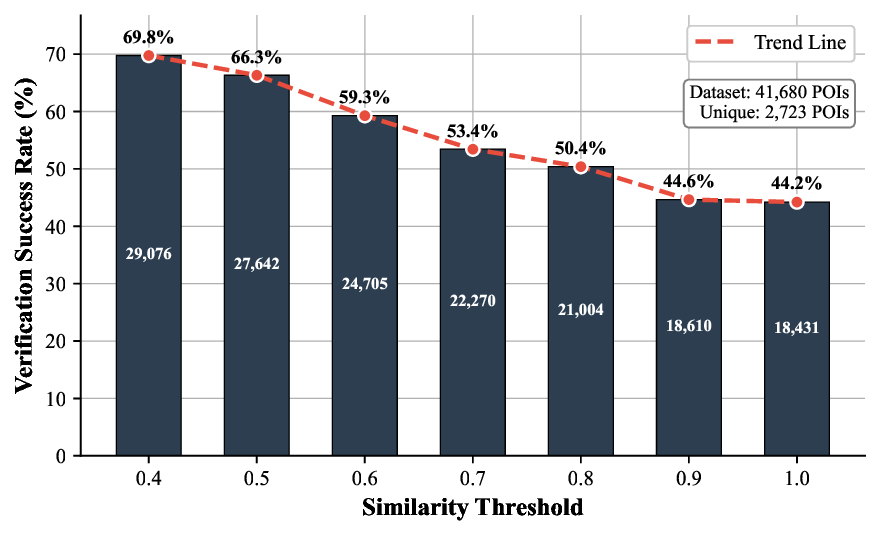}
        \caption*{(a) Beijing}
    \end{minipage}
    \hfill
    \begin{minipage}{0.48\columnwidth}
        \centering
        \includegraphics[width=\linewidth]{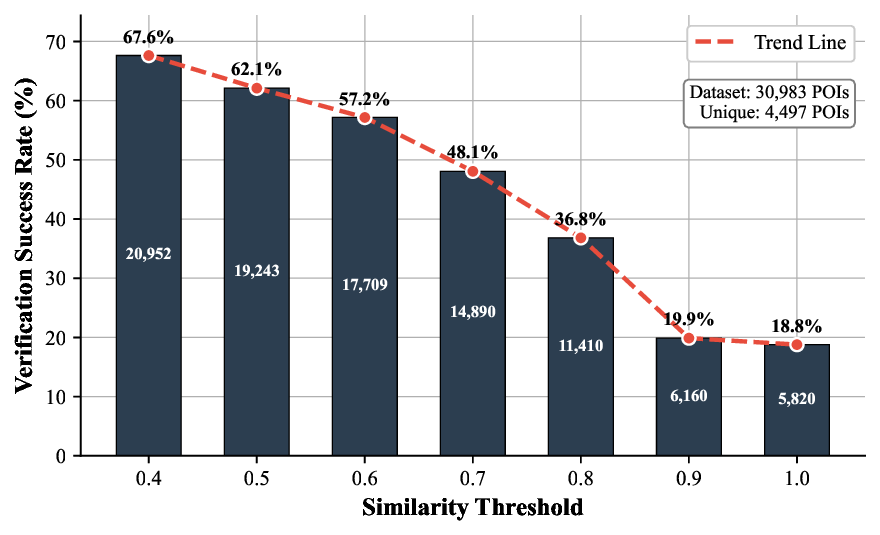}
        \caption*{(b) Tianjin}
    \end{minipage}
    \caption{Ratio of Real-World POIs in DeepSeek-R1 Generated Results on Beijing and Tianjin Datasets under Varying Similarity Thresholds}
    \label{fig:llm_flase}
\end{figure}

As outlined in Section~\ref{sec-exp-setup}, we first compared our approach with representative LLM-based baselines. 
The experimental results, presented in Table~\ref{tab:combined-data}, reveal that the LLM baselines frequently failed to meet user-specified travel days, indicating limited controllability. Additionally, its average total itinerary score was lower, primarily due to its insufficient capability in accurately filling schedules with appropriate POIs when users did not specify them. The values in parentheses in the $S_{\text{avg}}$ column represent the estimated theoretical maximums. These are computed by filling in all POIs that could not be successfully matched with the database using the average value of those that could be matched. This estimation method can be validated through LLM hallucination experiments, as shown in Fig.~\ref{fig:llm_flase}. When evaluating DeepSeek-R1 -- the best-performing model among LLM-only approaches -- its outputs exhibit a validation success rate of less than 70\%, even under a relatively low similarity threshold, indicating a significant number of hallucinated POIs. Under reasonable similarity thresholds, the success rate drops below 50\%. In contrast, UDOIP guarantees that 100\% of the generated POIs correspond to real-world locations, which further highlights its advantages in terms of controllability and feasibility. 
Moreover, the PTS method, which uses a traditional solver and focuses on POI types and quantity constraints, successfully eliminated or reduced day-count errors ($R_{days}$) and achieved a higher $S_{avg}$ than LLM baselines. However, its performance on preference alignment metrics ($H_1$ to $H_5$) is near zero. This indicates that while PTS can fill a schedule with the required number of POIs, it fails to optimize for route sequence, temporal habits, and nuanced user preferences.

\begin{figure}[H]\centering
    \includegraphics[width=1.0\columnwidth]{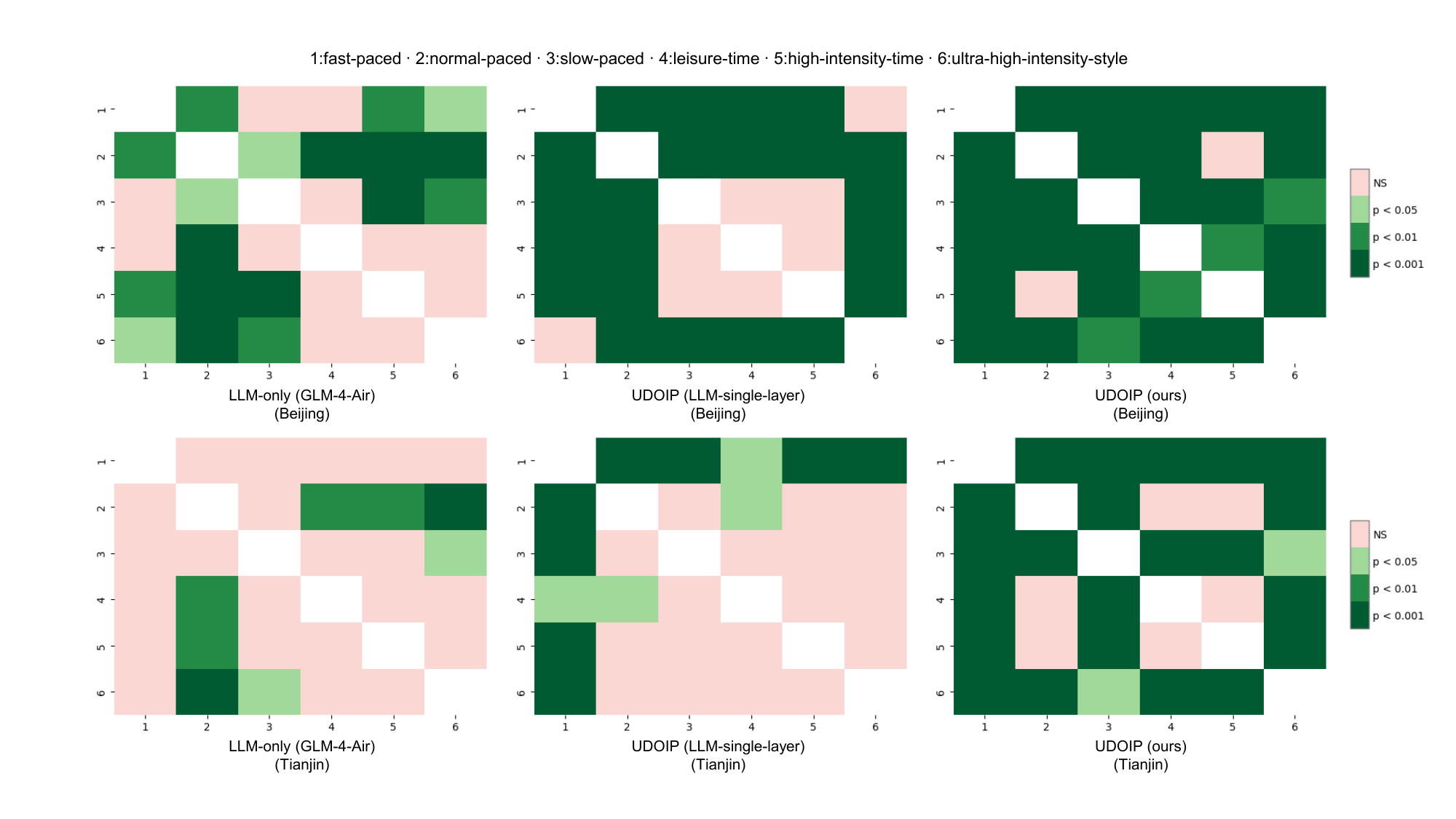}
    \captionsetup{justification=centering} 
    \caption{Heatmaps of P-values (\(p\text{-value}_{\text{Dunn}}\) under \(t^{visit}_p/t^{visit}_{p,u}\)) for Beijing and Tianjin Datasets (All Comparisons Are Conducted Using the GLM-4-Air Model)\\
    }
    \label{fig:method_dunn_p_4}
\end{figure}

Regarding preference extraction, our proposed LLM-based approach demonstrated higher test feature values in the preference variability test, which are presented in Table~\ref{tab:combined-data}. While our method achieved the best performance across most dimensions, discerning performance distinctions becomes challenging when multiple methods maintain consistently high values. To address this, we further analyzed inter-category preference differences using Dunn's test, as illustrated in Fig.~\ref{fig:method_dunn_p_4} and Fig.~\ref{fig:method_deepseek_dunn_p_4}\footnote{More relevant results can be found in Appendix
~\ref{ap-full-results}.\label{fn:ap-f}}. This indicates that UDOIP more effectively highlights the variability in results across different preference categories. Both tests were 
employed to assess the statistical significance of differences among these categories, and the higher values observed in our approach confirm its ability to capture and differentiate between distinct preference patterns with greater clarity.
Moreover, as shown in Fig.~\ref{fig:method_deepseek_dunn_p_4},
even advanced LLMs can only distinguish between trip categories with obvious semantic features, while ignoring many potential or less apparent feature requirements in user expressions. This is not what users expect to see.

\begin{figure}[H]\centering
    \includegraphics[width=1.0\columnwidth]{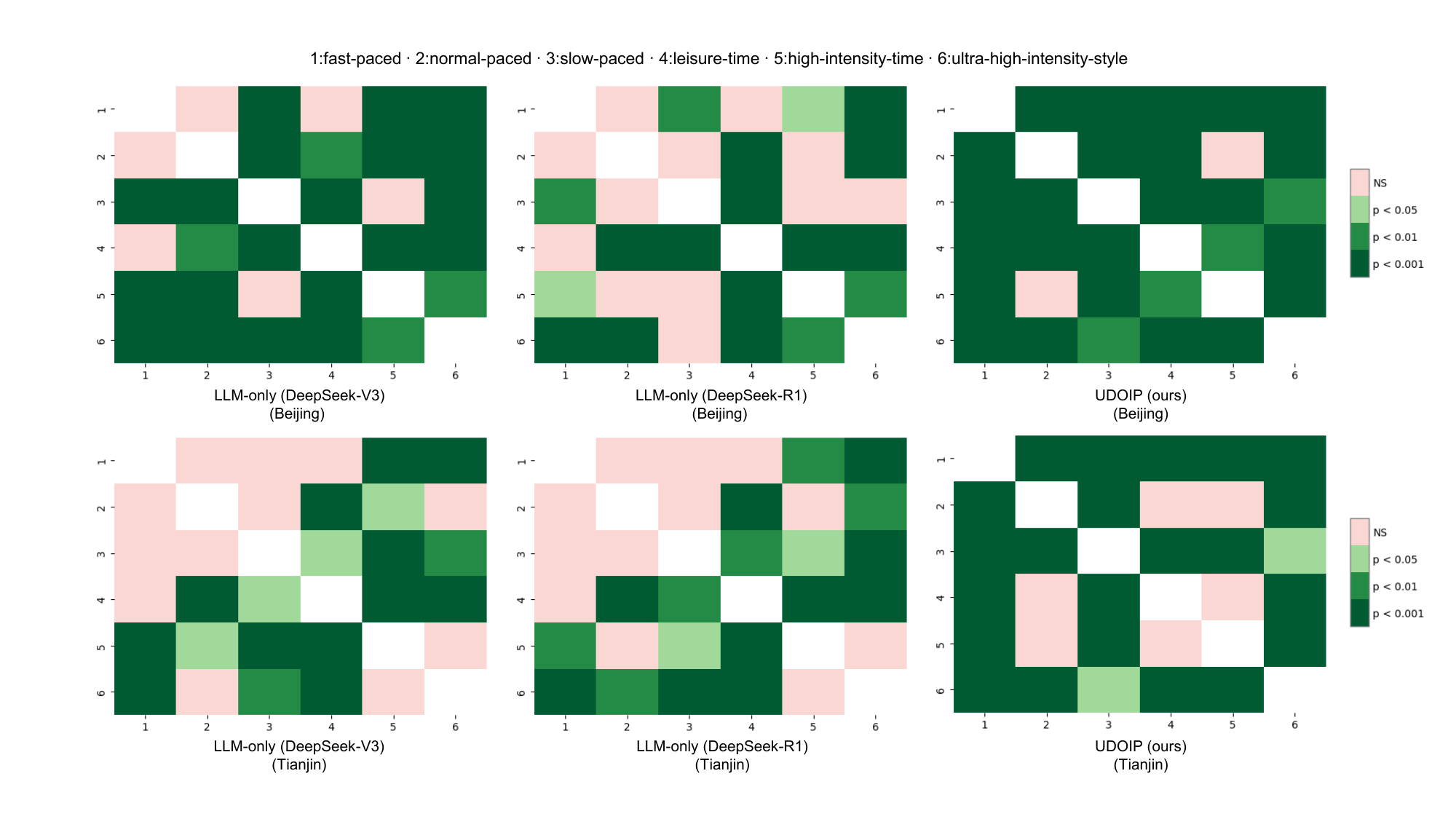}
    \captionsetup{justification=centering} 
    \caption{Heatmaps of P-values (\(p\text{-value}_{\text{Dunn}}\) under \(t^{visit}_p/t^{visit}_{p,u}\)) for Beijing and Tianjin Datasets (Comparing with DeepSeek Models)\\
    }
    \label{fig:method_deepseek_dunn_p_4}
\end{figure}

Additionally, we incorporated a second LLM classification layer to fine-tune incorrect inputs in our proposed method. When this second layer was removed, the result is shown in Table~\ref{tab:combined-data}. For instance, regarding time compactness, as shown in Fig.~\ref{fig:method_dunn_p_4}, the single-layer method struggled to distinguish between slow-paced and highly time-intensive preferences, which should have shown clear differences actually.
Although the complete UDOIP method cannot fully distinguish between similar concept pairs such as ``3: slow-pace'' and ``4: leisure-time'', it outperforms the other two methods entirely when applied to type groups with significant differences.

\subsection{Ablation Experiment (RQ3)}
\label{exp-ablation}

\begin{table*}[h]
\centering
\caption{Performance Metrics Comparison for LLM-FGRASP Variants in Beijing and Tianjin}
\label{tab:performance-metrics}
\resizebox{\textwidth}{!}{
    \begin{tabular}{ccccccc}
    \toprule
    \multicolumn{7}{c}{\textbf{Beijing}} \\
    \midrule
          & $S_{\text{avg}}$ $\uparrow$ & $D_{\text{total}}$ $\downarrow$ & $D_{\text{repeated}}$ $\downarrow$ & $\theta_{\text{turns}}$ $\downarrow$ & $A_{\text{overlap}}$ $\downarrow$ & $T_{\text{iter}}$ (s) $\downarrow$ \\
    \cmidrule{2-7}
    $\bullet$ LLM-FGR, 1Iter & 194.59  & 0.3   & 0.0488 & 1.91  & 0.002254  & 1.11  \\
    LLM-FGR, 2Iter & 197.51  & 0.3   & 0.0492 & 1.91  & 0.002241  & 2.15  \\
    LLM-FGR+C, 1Iter & 202.59  & 0.28  & 0.048 & 1.95  & 0.001044  & 1.30  \\
    $\blacktriangle$ LLM-FGR+C, 2Iter & \underline{206.76}  & 0.28  & 0.0493 & 1.94  & \textbf{0.000996 } & 1.53  \\
    LLM-FGR+CS (ours), 1Iter & \textbf{210.15 } & \textbf{0.25} & \textbf{0.0452} & \textbf{1.86} & 0.001012  & \textbf{1.05 } \\
    $\bullet$ Improvement-original & 8.00\% & 16.67\% & 7.38\% & 2.62\% & 55.12\% & 5.41\% \\
    $\blacktriangle$ Improvement-suboptimal & 1.64\% & 10.71\% & 8.32\% & 4.12\% & -1.58\% & 31.37\% \\
    \midrule
    \multicolumn{7}{c}{\textbf{Tianjin}} \\
    \midrule
          & $S_{\text{avg}}$ $\uparrow$ & $D_{\text{total}}$ $\downarrow$ & $D_{\text{repeated}}$ $\downarrow$ & $\theta_{\text{turns}}$ $\downarrow$ & $A_{\text{overlap}}$ $\downarrow$ & $T_{\text{iter}}$ (s) $\downarrow$ \\
    \cmidrule{2-7}
    $\bullet$ LLM-FGR, 1Iter & 230.06  & 0.64  & 0.2196 & 1.86  & 0.002178  & 1.03  \\
    $\blacktriangle$ LLM-FGR, 2Iter & \textbf{233.70 } & 0.63  & 0.217 & 1.86  & 0.002257  & 1.87  \\
    LLM-FGR+C, 1Iter & 226.26  & 0.56  & 0.1628 & 1.9   & 0.000792  & 1.28  \\
    LLM-FGR+C, 2Iter & 229.96  & 0.56  & 0.1621 & 1.9   & 0.000879  & 1.52  \\
    LLM-FGR+CS (ours), 1Iter & \underline{231.58}  & \textbf{0.52} & \textbf{0.1344} & \textbf{1.84} & \textbf{0.000631 } & \textbf{1.02 } \\
    $\bullet$ Improvement-original & 0.66\% & 18.75\% & 38.80\% & 1.08\% & 71.05\% & 0.97\% \\
    $\blacktriangle$ Improvement-suboptimal & -0.91\% & 17.46\% & 38.06\% & 1.08\% & 72.06\% & 45.45\% \\
    \bottomrule
    \end{tabular}%
}
\parbox{1.0\linewidth}{\footnotesize{Note: LLM-FGR refers to the LLM-based FGRASP method; ``+C'' 
denotes the addition of clustering; ``+CS'' 
signifies the inclusion of both clustering and substitution; ``Iter'' 
stands for iterations; ``Improvement'' 
indicates improvement over the original method or the suboptimal approach measuring by $S_{avg}$.}}
\end{table*}


The results of ablation study in Table~\ref{tab:performance-metrics} compare the performance metrics of various LLM-FGRASP model variants, including the basic LLM-FGR with one and two iterations, LLM-FGR+C (clustering addition) with one and two iterations, and LLM-FGR+CS, which incorporates both clustering and substitution, also referred to as the complete UDOIP framework. Each variant introduces specific modifications aimed at improving the planning efficiency within a short time frame, reducing overlapping areas in multi-day itineraries, and minimizing detours in single-day itineraries.

In particular, the clustering strategy (denoted as ``C''  in LLM-FGR+C) shows notable improvements in $A_{\text{overlap}}$ by grouping POIs that are spatially or temporally closer, which minimizes redundant travel. However, in the case of Tianjin, the improvement on $S_{\text{avg}}$ remains minimal, and some scores even decline, suggesting that clustering alone may not fully optimize the routing in this specific context. The inclusion of substitution (denoted as ``CS''  in LLM-FGR+CS) further enhances performance by allowing the replacement of less relevant POIs with alternatives that better satisfy user constraints. Notably, the complete UDOIP framework (LLM-FGR+CS) achieves the highest computational efficiency, yielding the shortest $T_{iter}$ while providing comprehensive metric improvements within a single iteration, outperforming multi-iteration baselines. 

To interpret the performance gains, we provide two types of comparisons: one with the original results and another with suboptimal results. The improvement over the original results is assessed relative to the initial configuration without clustering and substitution. For the suboptimal comparison, we measure against the highest \( S_{\text{avg}} \) score across all other models, rather than selecting the second-best values for each individual metric. This approach ensures consistency by comparing with a single suboptimal baseline rather than varying targets across metrics.

As indicated in Table~\ref{tab:performance-metrics}, the LLM-FGR+CS (UDOIP) model achieves the most significant improvements across nearly all metrics in Beijing. In Tianjin, while its $S_{\text{avg}}$ is second to the 2-iteration basic variant, it reaches the best values in all spatial constraints ($D_{\text{total}}$, $A_{\text{overlap}}$, etc.) with the lowest computational cost. This demonstrates the model's effectiveness in balancing computational efficiency and solution feasibility, thereby yielding routes that are not only efficient but also practically implementable.



\subsection{Case Study (RQ4 RQ5)}
\label{sec-exp-case}

\begin{figure*}[h]\centering
    \includegraphics[width=1.0\columnwidth]{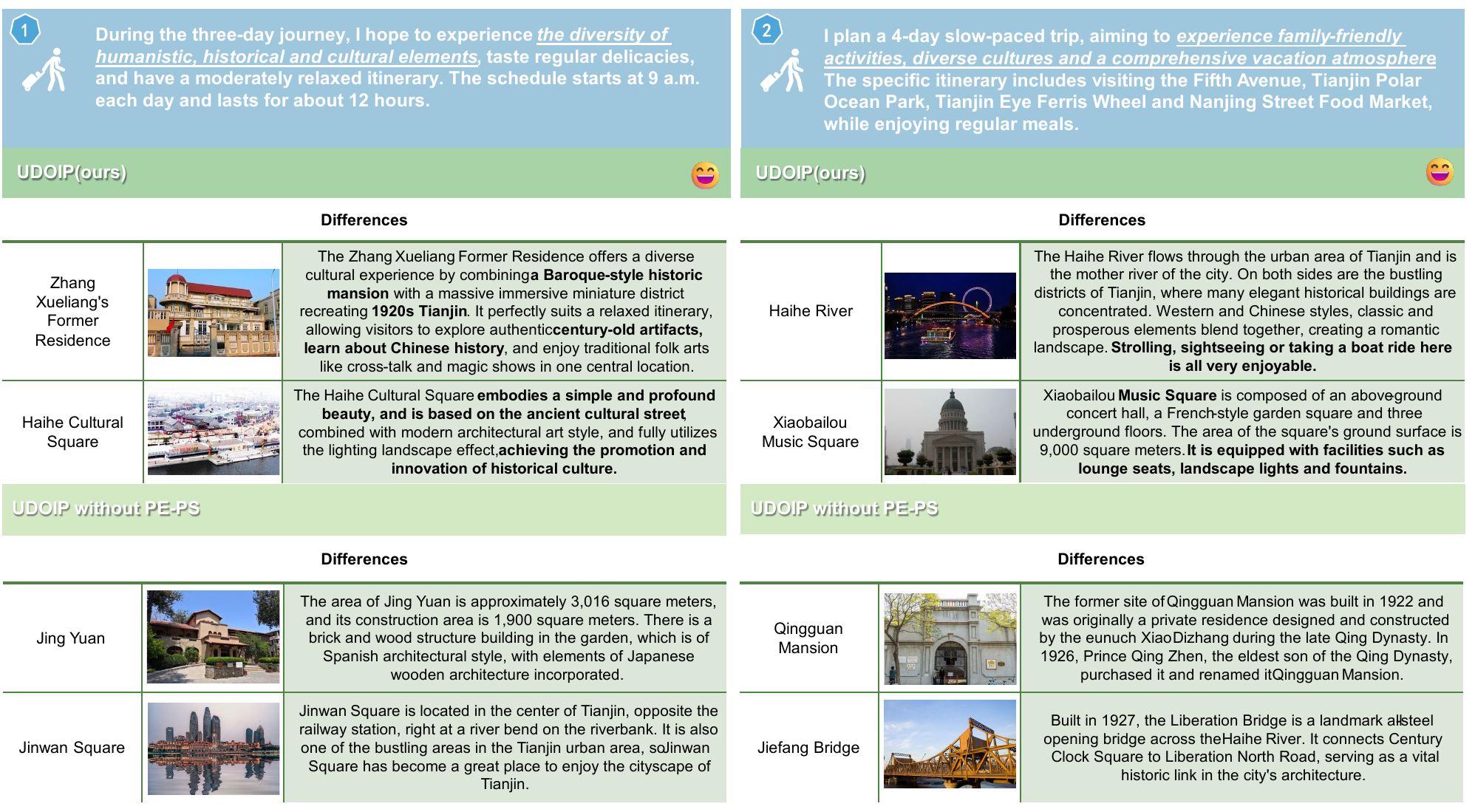}
    \captionsetup{justification=centering} 
    \caption{Comparison of Two Users' Itinerary Planning: UDOIP Versus UDOIP without PE-PS}
    \label{fig:case_study_0}
\end{figure*}

We have introduced 
PE-PS in section~\ref{opt-grasp} to integrate personalized user preferences into traditional planning algorithms. To test the effectiveness of using 
this method to align with user preferences, we compared the results of the complete model with those of a model excluding PE-PS to address the differences under two distinct user requirements, as shown in Fig.~\ref{fig:case_study_0}. 

For the user requirements on the left, the complete UDOIP's recommendations of Zhang Xueliang's Former Residence and Haihe Cultural Square better align with the user's interests in humanistic and historical diversity. Zhang Xueliang's Former Residence offers a profound exploration of 1920s Tianjin and Chinese history through its artifacts and traditional folk arts, which is more suited to the user's request for cultural depth compared to Jing Yuan, which focuses more on architectural descriptions. Similarly, Haihe Cultural Square emphasizes the heritage of ancient cultural streets and historical innovation, providing a stronger sense of cultural immersion than the more visually-oriented Jinwan Square.

The user preferences illustrated on the right emphasize family-friendly activities and a comprehensive vacation atmosphere. In the complete UDOIP model, Haihe River and Xiaobailou Music Square are selected as representative POIs. The Haihe River offers sightseeing boat rides and strolling experiences that are ideal for family-friendly vacations, while Xiaobailou Music Square provides a relaxed atmosphere with its fountains, landscape lights, and lounge seats. In contrast, after removing the PE-PS module, these were replaced by Qingguan Mansion, which focuses on private residence history, and Jiefang Bridge, a landmark architectural link, both of which align less closely with the user's desire for a family-oriented vacation vibe.

This demonstrates that using PE-PS allows for a more effective alignment with user preferences, addressing the limitations of relying solely on fixed scores to retrieve POIs.

\begin{figure}[h]\centering
    \includegraphics[width=1.0\columnwidth]{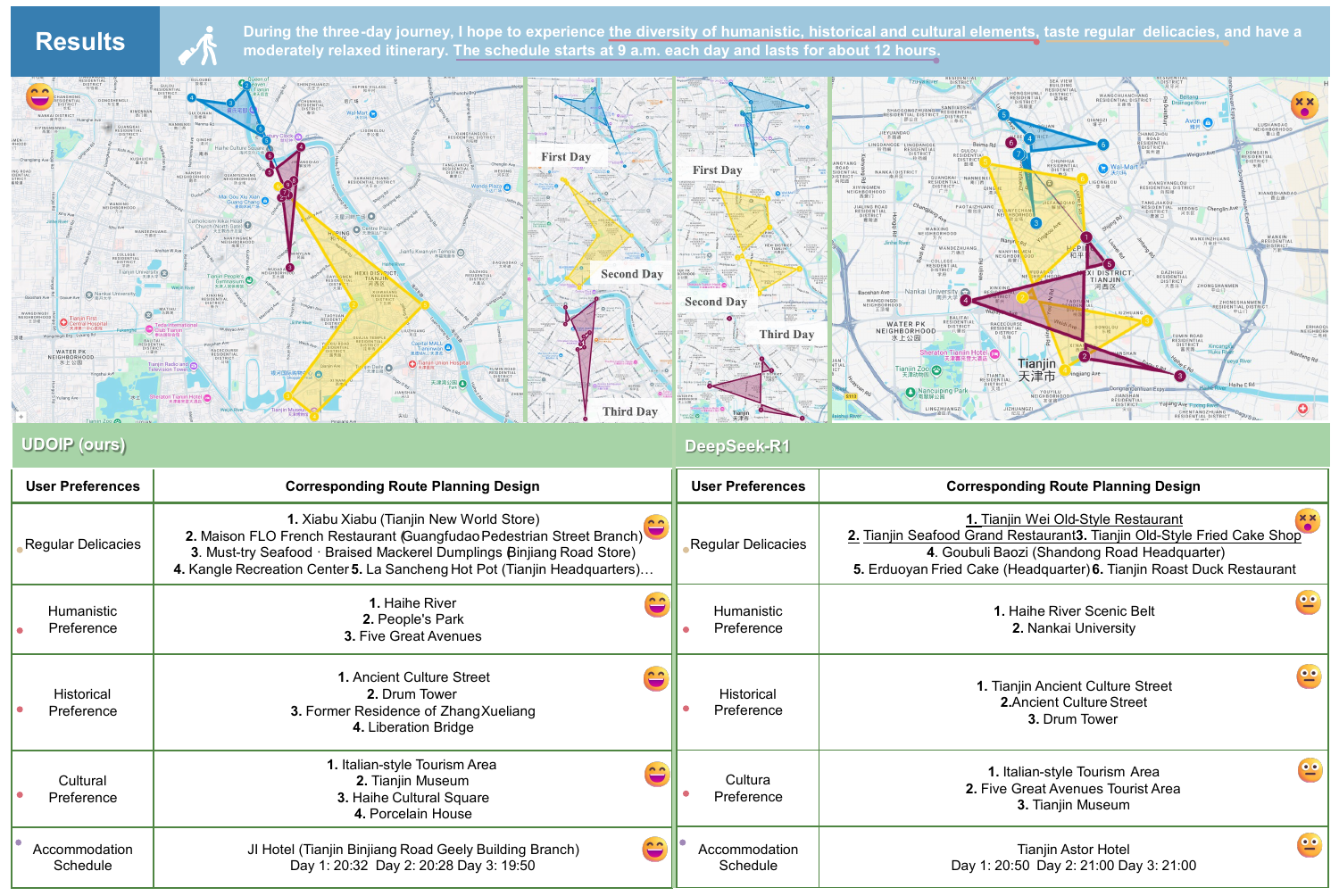}
    \captionsetup{justification=centering} 
    \caption{Comparison of Three-Day Itinerary Planning: UDOIP Versus Deepseek-R1 Results}
    \label{fig:case_study_1}
\end{figure}

In addition, to demonstrate the clear advantages of our approach, we conducted comparative experiments using real-world user requirements as inputs for both our method and DeepSeek-R1, which demonstrated the best performance in prior experiments.
The outputs of both models were visualized on Google Maps\footnote{http://www.google.com/maps/} to mark the POIs and routes, providing a direct comparison of the results, as shown in Fig.~\ref{fig:case_study_1}.

As illustrated in Fig.~\ref{fig:case_study_1}, we compare the planning results of UDOIP and DeepSeek-R1 for a three-day itinerary. In terms of spatial organization, UDOIP exhibits superior geographical awareness; its multi-day routes show distinct regional clusters (First Day in blue, Second Day in yellow, and Third Day in purple), with minimal overlapping or backtracking within each single-day itinerary. In contrast, DeepSeek-R1 demonstrates limited spatial awareness, resulting in overlapping geographic areas across different days and significant detours, which increases unnecessary travel time for users.

Regarding generation accuracy, UDOIP ensures 100\% POI feasibility by retrieving data from a validated real-world database. However, DeepSeek-R1 suffers from significant "hallucinations." As indicated by the underlined entries in the comparison table, it generated three fabricated -- spurious entries that render the generated itinerary unusable in practice.

Furthermore, in terms of scheduling, UDOIP provides much higher precision, accurately calculates travel times between POIs and provides specific accommodation arrival times (e.g., 20:32, 20:28). Conversely, DeepSeek-R1 provides vague text-based outputs based solely on user descriptions, failing to account for actual traffic conditions or venue opening hours, resulting in a less realistic schedule (e.g., simplified 21:00 deadlines).

Lastly, in terms of computational efficiency, DeepSeek-R1 required an average of 44.92 seconds over ten runs to generate results, whereas UDOIP completed the task in just 3.63 seconds on average across the same ten executions, demonstrating significantly faster performance.
At the same time, our framework only relies on a small set of parameters from the LLM to achieve highly controllable results, making it more cost-effective while ensuring greater reliability, which makes it well-suited for commercial use.

\section{Conclusion}

This paper studies multi-day urban travel itinerary planning (MUTIP) in real-world settings, where solutions must jointly satisfy spatiotemporal constraints and diverse user preferences, while remaining practical for interactive use. We present UDOIP, a hybrid framework that uses a dual-layer LLM pipeline to translate natural-language requests into structured algorithm inputs, and a preference-aware heuristic planner to produce feasible multi-day itineraries. Experiments on two real-world city datasets and a case study with DeepSeek-R1 show that UDOIP improves controllability, preference alignment, and planning efficiency compared with representative LLM-based baselines. In future work, we plan to extend the framework to broader domains and incorporate more flexible preference sources and open-domain data.

\printbibliography

\appendix
\newpage

\section{Algorithms}
\label{ap:alg}
\renewcommand{\thealgorithm}{\Alph{section}.\arabic{algorithm}} 
\setcounter{algorithm}{0} 

In this section, we provide the complete pseudocode for the method described in Sections \ref{method-llms}-\ref{opt-grasp}. 

\begin{breakablealgorithm}
\caption{LLMs-based Input Parameters Generation Framework}
\label{alg:llm-input-gen}
\begin{algorithmic}[1]
\Require User natural language description
\Ensure input parameters relevant to user preferences
\State \textbf{Layer 1: Overall Assignment Layer}
\State Generate initial input parameters from user description using Layer 1 LLM
\Statex \quad \textit{(e.g., prioritized POI list, dining preferences, time settings)}
\State \textbf{Layer 2: Classification Fine-tuning Layer}
\State Classify user preferences into categories that exhibit distinct preferences.
\Statex \quad \textit{(e.g., ``fast-paced'', ``normal-paced'', ``slow-paced'')}
\State Retrieve predefined acceptable ranges for each category
\If {any parameter values are outside the typical range}
    \State Adjust parameters to fall within the acceptable range
\EndIf
\State Return the final input parameters
\end{algorithmic}
\end{breakablealgorithm}

\begin{breakablealgorithm}
\caption{Itinerary Construct Phase}
\label{alg:rcl-construct}
\begin{algorithmic}[1]
\Require POI set retrieved from the database, user preferences, available time, size of RCL $RCL\_SIZE$, size of RCL* ${RCL}^{*}\_SIZE$, current itinerary $I$
\Ensure Updated itinerary $I$

\If{$I$ is empty}
    \State Initialize $I$ by appending a hotel-type POI twice per day as both the starting and ending point of the itinerary.
\EndIf
\State Initialize \(LOCK\) = \textbf{True}
\State Compute embeddings of POIs using Formula (\ref{eq:emb_poi})
\State Compute embedding of user using Formula (\ref{eq:average_emb})
\While{The POI set is not empty, and there is sufficient remaining time to continue adding POIs to $I$}
\State Initialize RCL and RCL* as empty lists
\For{each POI \( p \) in the POI set}
    \If{\(LOCK\) and \textbf{POI does not cater to dining needs or preferred POIs}}
        \State continue
    \EndIf
    \State Choose the optimal insertion position using a greedy time-based strategy, where the corresponding increase in travel time is calculated as \(t^{traffic}_{(p, p_{+1})} + t^{traffic}_{(p_{-1}, p)} - t^{traffic}_{(p_{-1}, p_{+1})} + t^{visit}_{(p,u)}\), where $p_{-1}$ and $p_{+1}$ are the POIs immediately before and after the simulated insertion position within the current itinerary $I$
        \If{travel time meets $t_u^{plan}$ and within time window}
            \State Add \( (p, \text{optimal insertion position for }p) \) to RCL
        \EndIf
\EndFor
\If{RCL is empty for two consecutive iterations}
    \State Set \(LOCK\) = \textbf{False} and jump to Line 7
\EndIf
\State Sort the RCL in ascending order based on the increase in travel time, then select the top $RCL\_SIZE$ size of RCL elements.
\State Calculate similarity \( sim_{(p,u)} \) using Formula (\ref{eq:similarity}) for all POI in RCL
\State Compute score \( s_{(p,u)} \) using Formula (\ref{eq:score}) for all POI in RCL
\State Construct the RCL* by sorting the POIs according to their scores and selecting the top ${RCL}^{*}\_SIZE$ elements.
\State Randomly select a tuple 
\( (p, \text{optimal insertion position for }p) \) from the RCL* and insert the POI $p$ into $I$ at the corresponding optimal insertion position.
\EndWhile
\State Return $I$
\end{algorithmic}
\end{breakablealgorithm}

\begin{breakablealgorithm}
\caption{Local Search Phase}
\label{alg:local-search}
\begin{algorithmic}[1]
\Require Current best, POI time window constraints
\Ensure Improved itinerary with waiting time, optimal time window, and travel time
\State Evaluate current best itinerary using Formula (\ref{eq:evaluation}) as current best result of optimization function
\State Initialize $FLAG_{improvement}$ = \textbf{True}
\While{$FLAG_{improvement}$ is \textbf{True}}
    \State Set $FLAG_{improvement}$ = \textbf{False}
    \For{each day $k$ in the itinerary}
        \For{each sub-sequence $i,j$ where $1 \leq i < j < len(I[k]) - 1$}
            \State Create new itinerary by reversing the sub-sequence from $i$ to $j$
            \If{new itinerary meets $t^{window_{open}}_p$ of all POIs}
                \State Evaluate new itinerary using Formula (\ref{eq:evaluation})
                \If{new itinerary has smaller result of optimization objective function}
                    \State Update current best itinerary and best result of optimization function
                    \State Set $FLAG_{improvement}$ = \textbf{True}
                \EndIf    
            \EndIf
        \EndFor
    \EndFor
\EndWhile
\State Return local best itinerary solution
\end{algorithmic}
\end{breakablealgorithm}

\begin{breakablealgorithm}
\caption{Clustering and Substitution}
\label{alg:clustering-substitution}
\begin{algorithmic}[1]
\Require POI set retrieved from the database, user preferences, total number of days \( k \), time threshold \( \gamma \)
\Ensure Optimized itinerary with clusters
\State \textbf{Cluster POIs into \( k \) groups based on geographic locations}
\For{each cluster day}
    \Statex \quad \textit{(Plan route within the cluster by Algorithm~\ref{alg:rcl-construct} and \ref{alg:local-search})}
    \If{\textbf{POIs in the cluster do not meet all user preferences}}
        \State Check cross-cluster POIs for substitutions
        \If{substitution meets user constraints or exceeds time saving threshold \( \gamma \)}
            \State Replace current POI with substitute POI from another cluster
        \EndIf
    \EndIf
\EndFor
\State Return the itinerary optimized by clustering and substitution
\end{algorithmic}
\end{breakablealgorithm}

\section{Prompts in UDOIP}
\label{ap-prompts}

In this section, we present the key prompts used in our proposed methodology. The prompts are intended to clearly outline the controlled variables of the UDOIP, ensuring reproducibility of our results.

\subsection{Overall Assignment Layer}
\begin{tcolorbox}[colback=gray!5!white, colframe=gray!75!black, title=Overall Assignment Layer]
You serve as a translator between human language and itinerary planning algorithms, capable of accurately interpreting and extracting users' travel preferences to correspond with algorithm input parameters. Your job is to understand user requirements in a single step based on your existing knowledge (without seeking additional information) and provide appropriate function input parameters based on a single interaction. The user's destination is \texttt{{\{city name\}}}, and all Points of Interest (POIs) must be listed with complete and precise information in Chinese, without abbreviations. The output should not include any comments.

\texttt{{\{format\_instructions\}}}
\end{tcolorbox}

\subsection{Classification Fine-Tuning Layer}
\subsubsection{Time Preference}
\begin{tcolorbox}[colback=gray!5!white, colframe=gray!75!black, title=Time Preference Classifier]
You are a trip time preference classifier that categorizes users' travel time preferences into one of six categories based on their itinerary planning needs: fast-paced, normal-paced, slow-paced, leisure-time, high-intensity-time, and ultra-high-intensity-style.

\texttt{{\{format\_instructions\}}}
\end{tcolorbox}

\subsubsection{Dining Preference}
\begin{tcolorbox}[colback=gray!5!white, colframe=gray!75!black, title=Dining Preference Classifier]
You are a dining preference classifier that categorizes users' dining preferences into one of two categories based on their itinerary planning needs: regular (e.g., two or three meals a day, home-style dishes) and gourmet (e.g., food exploration, specialty cuisine).

\texttt{{\{format\_instructions\}}}
\end{tcolorbox}

\section{Datasets}
\label{ap-datasets}
In this section, we provide two types of datasets: POI data and simulated user demands data, intended for practical scenario analysis. The datasets are presented below in a card-like format.

\begin{tcolorbox}[colback=gray!5!white, colframe=gray!75!black, title=POI Dataset, label=fig:poi_data]
    \textbf{id:} 1 \\
    \textbf{name:} The Palace Museum \\
    \textbf{city:} Beijing \\
    \textbf{rating:} 5.0 \\
    \textbf{rating number:} 15xxxx \\
    \textbf{latitude:} 39.xxxxxx \\
    \textbf{longitude:} 116.xxxxxx \\
    \textbf{category:} scenic spot \\
    \textbf{suggested visiting duration:} 8h \\
    \textbf{address:} 4 Jingshanqian Street, Dongcheng District, Beijing \\
    \textbf{description:} The Palace Museum, also known as the Forbidden City, was the imperial palace of the Ming and Qing dynasties.
\end{tcolorbox}

\vspace{1em}

\begin{tcolorbox}[colback=gray!5!white, colframe=gray!75!black, title=Simulated User Demand Dataset, label=fig:user_demand_data]
    \textbf{id:} 1 \\
    \textbf{simulated user demand:} I'd like to plan a 6-day classic and well-rounded trip. I want to visit the Badaling Great Wall and the Water Cube. For food, I'd prefer something simple and typical. I want a relaxed pace, starting each day around 9 AM, with no more than 12 hours of activities per day. \\
    \textbf{prioritized POIs:} Badaling Great Wall, The Water Cube \\
    \textbf{time preference:} normal-paced \\
    \textbf{dining preference:} regular
\end{tcolorbox}

\section{Prompt in Experiments}
\label{ap-exp-prompts}

In the experimental section, we compared the UDOIP method with approaches that rely solely on LLMs. To ensure the validity and clarity of the comparison, we maintained consistency in the prompts as much as possible. The prompt for the LLM-only method is as follows.

\subsection{Prompt of LLM-only Method}
\begin{tcolorbox}[colback=gray!5!white, colframe=gray!75!black, title=LLM-only Method]
You serve as an itinerary planning assistant, capable of accurately interpreting and extracting users' travel preferences to provide specific planning results. Your job is to understand user requirements in a single step and give a reasonable daily travel plan based on their requirements. Each day must start from the hotel and end at the hotel, ensuring a sufficient number of restaurants and attractions that match the user's dining needs and time rhythm. The time, whether early or late, and the pace, whether fast or slow, must also meet the user's preferences. The distance between POIs and the reasonableness of the route should also be considered. The user's destination is \texttt{{\{city name\}}}, and all Points of Interest (POIs) must be listed with complete and precise information in Chinese, without abbreviations. The output should not include any comments.

\texttt{{\{format\_instructions\}}}
\end{tcolorbox}

\section{Full results of Dunn's test}
\label{ap-full-results}
\begin{figure*}[ht]
    \centering
    \begin{minipage}[b]{0.45\textwidth}
        \centering
        \includegraphics[width=\textwidth]{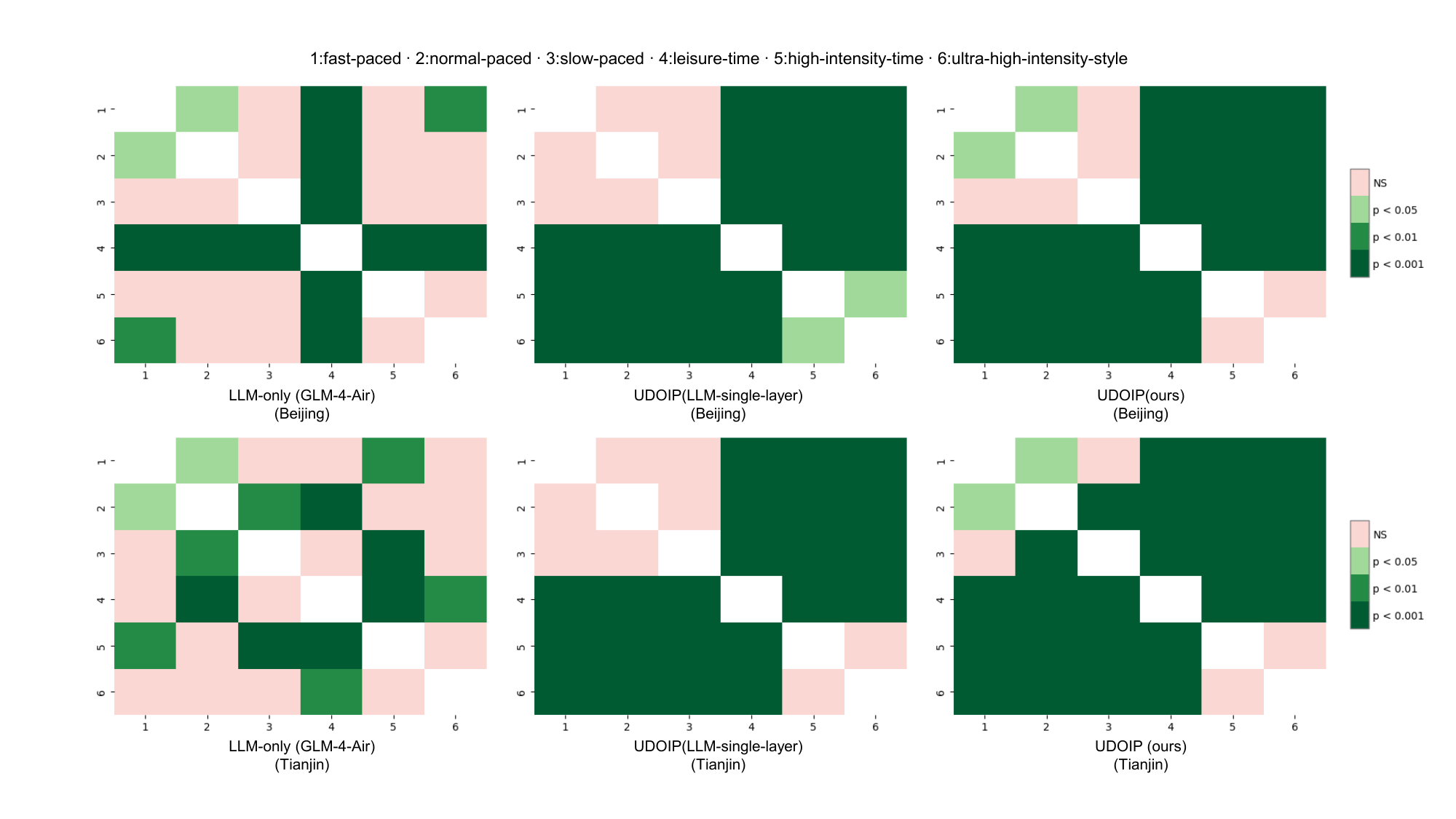}
        \caption{Average Daily Travel Time for Two Cities}
        \label{fig:avg_daily_travel_time}
    \end{minipage}
    \hfill
    \begin{minipage}[b]{0.45\textwidth}
        \centering
        \includegraphics[width=\textwidth]{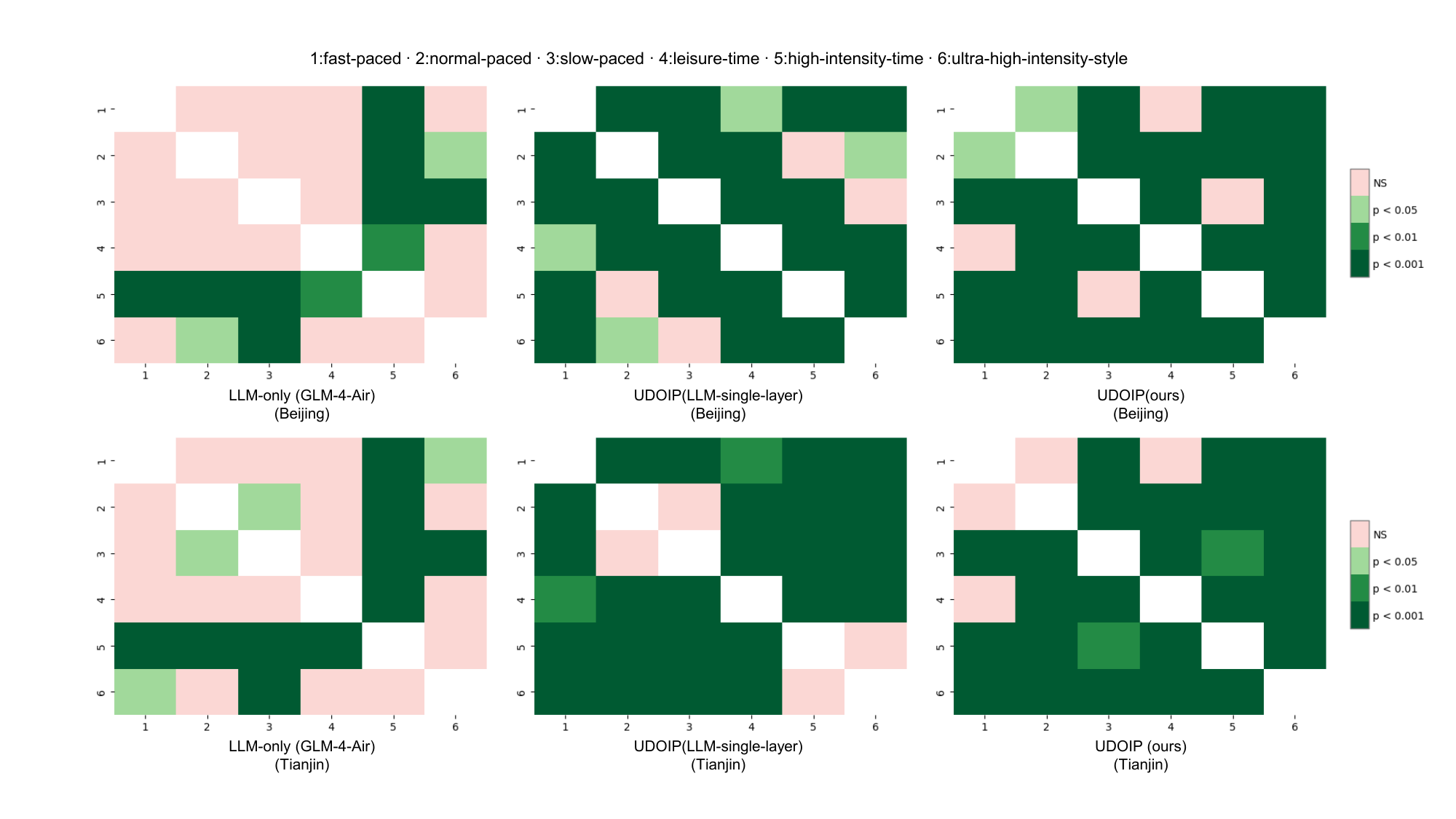}
        \caption{Average Original Recommendation Duration for Two Cities}
        \label{fig:avg_recommendation_duration}
    \end{minipage}
    \\
    \begin{minipage}[b]{0.45\textwidth}
        \centering
        \includegraphics[width=\textwidth]{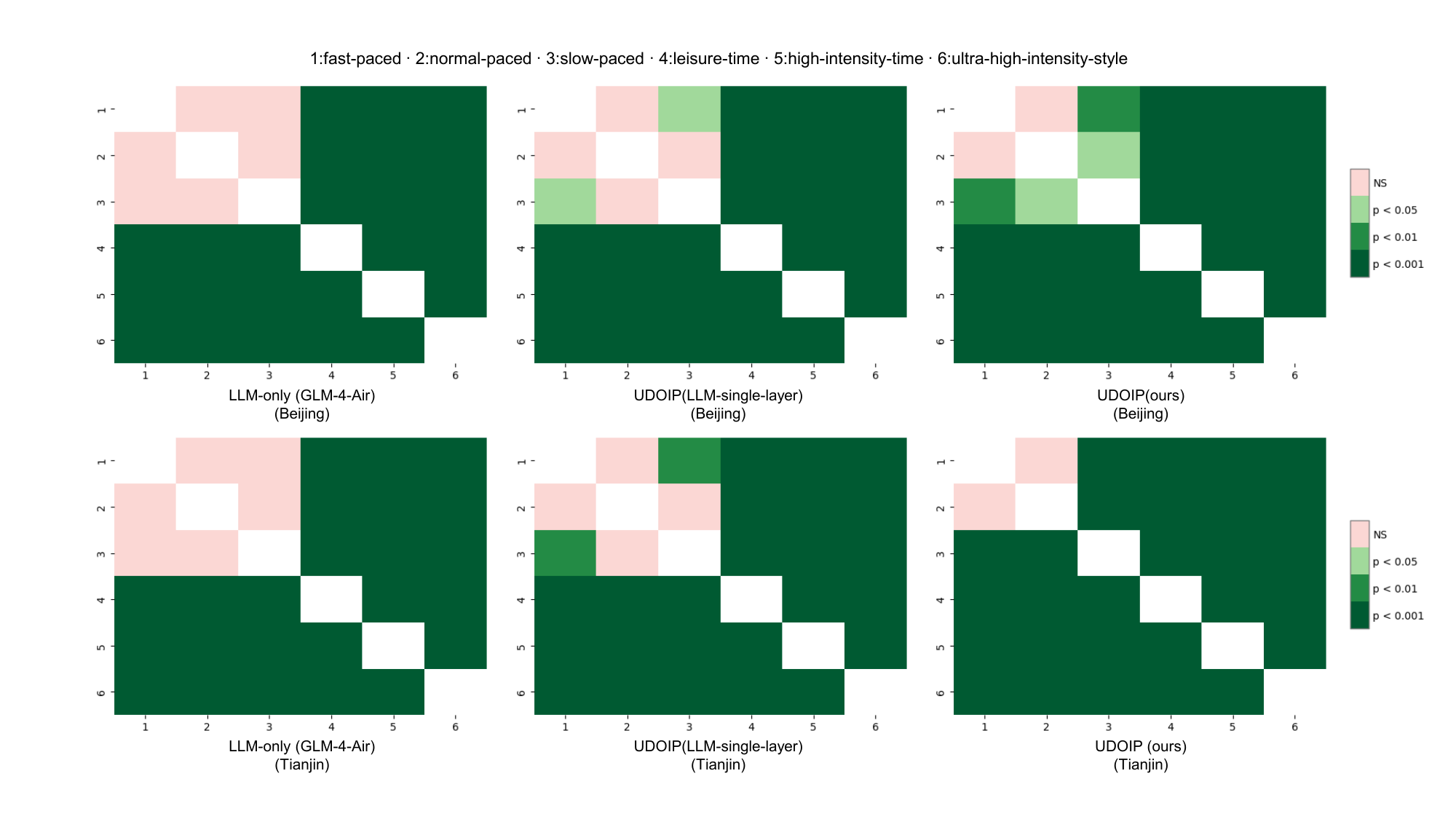}
        \caption{Average Daily Start Time for Two Cities}
        \label{fig:avg_start_time}
    \end{minipage}
    \hfill
    \begin{minipage}[b]{0.45\textwidth}
        \centering
        \includegraphics[width=\textwidth]{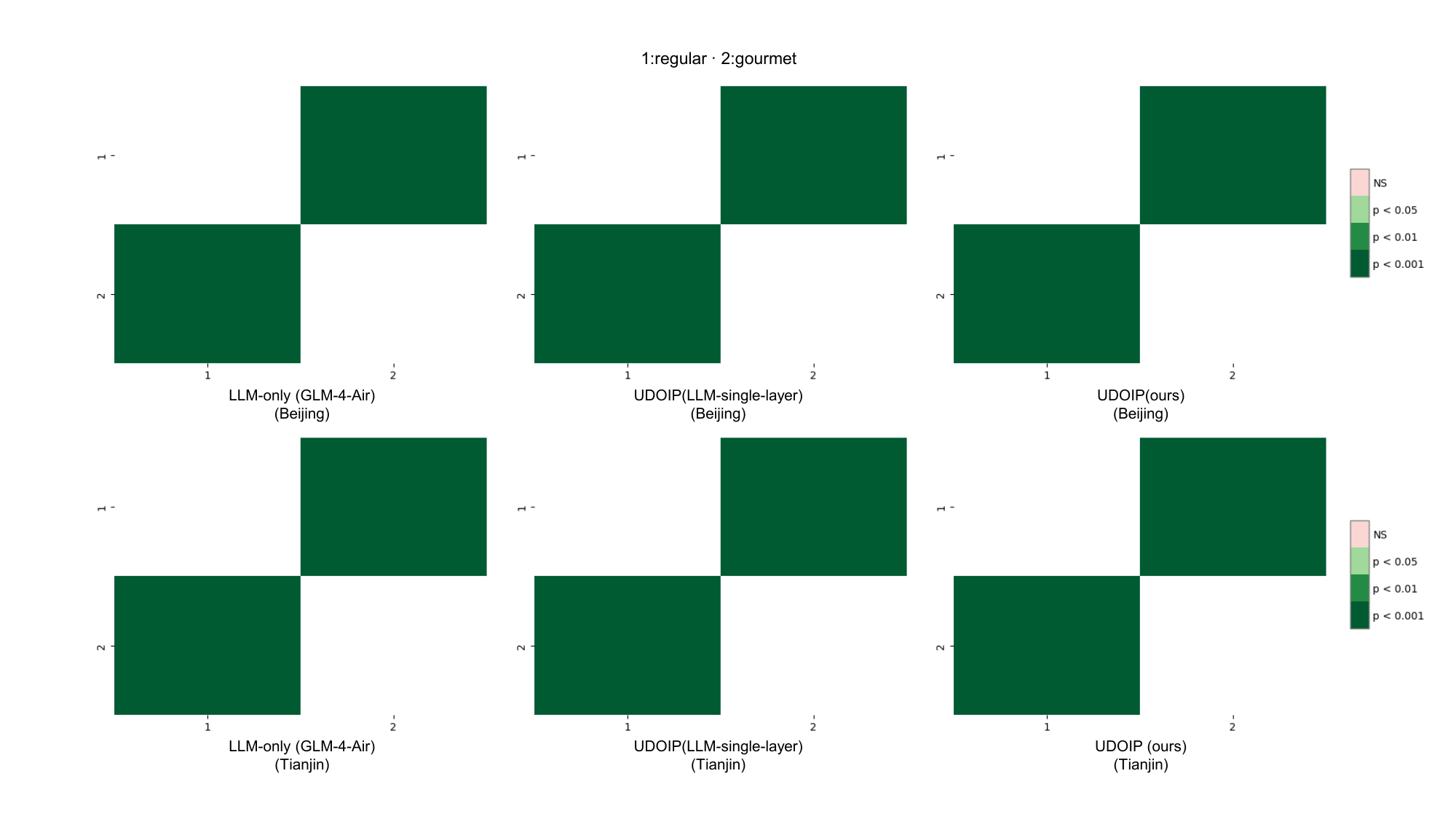}
        \caption{Total Number of Dining Experiences for Two Cities}
        \label{fig:total_dining_count}
    \end{minipage}
\end{figure*}

\begin{figure*}[ht]
    \centering
    \begin{minipage}[b]{0.45\textwidth}
        \centering
        \includegraphics[width=\textwidth]{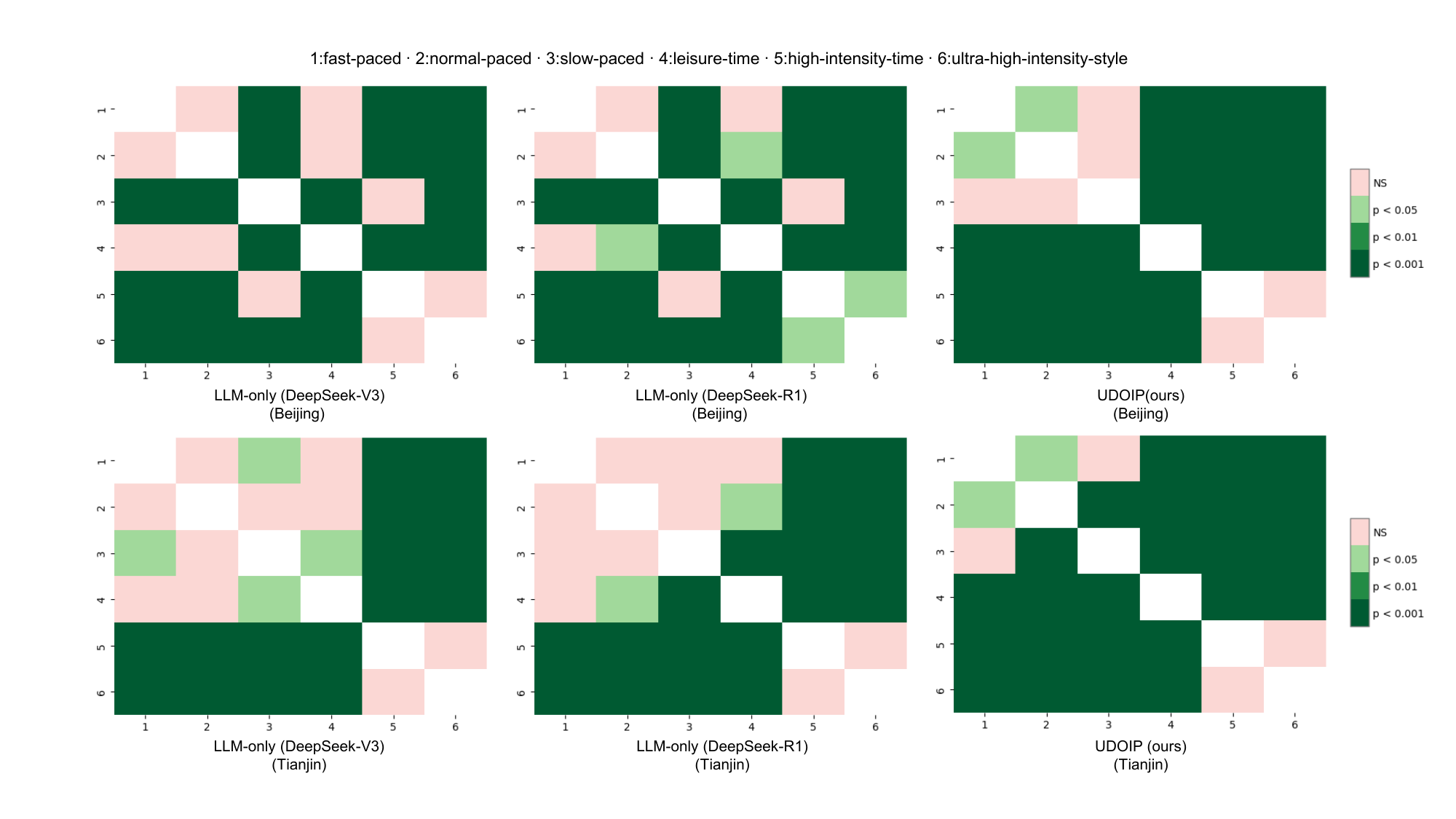}
        \caption{Average Daily Travel Time for Two Cities Compared with DeepSeek Model}
        \label{fig:avgd_daily_travel_time}
    \end{minipage}
    \hfill
    \begin{minipage}[b]{0.45\textwidth}
        \centering
        \includegraphics[width=\textwidth]{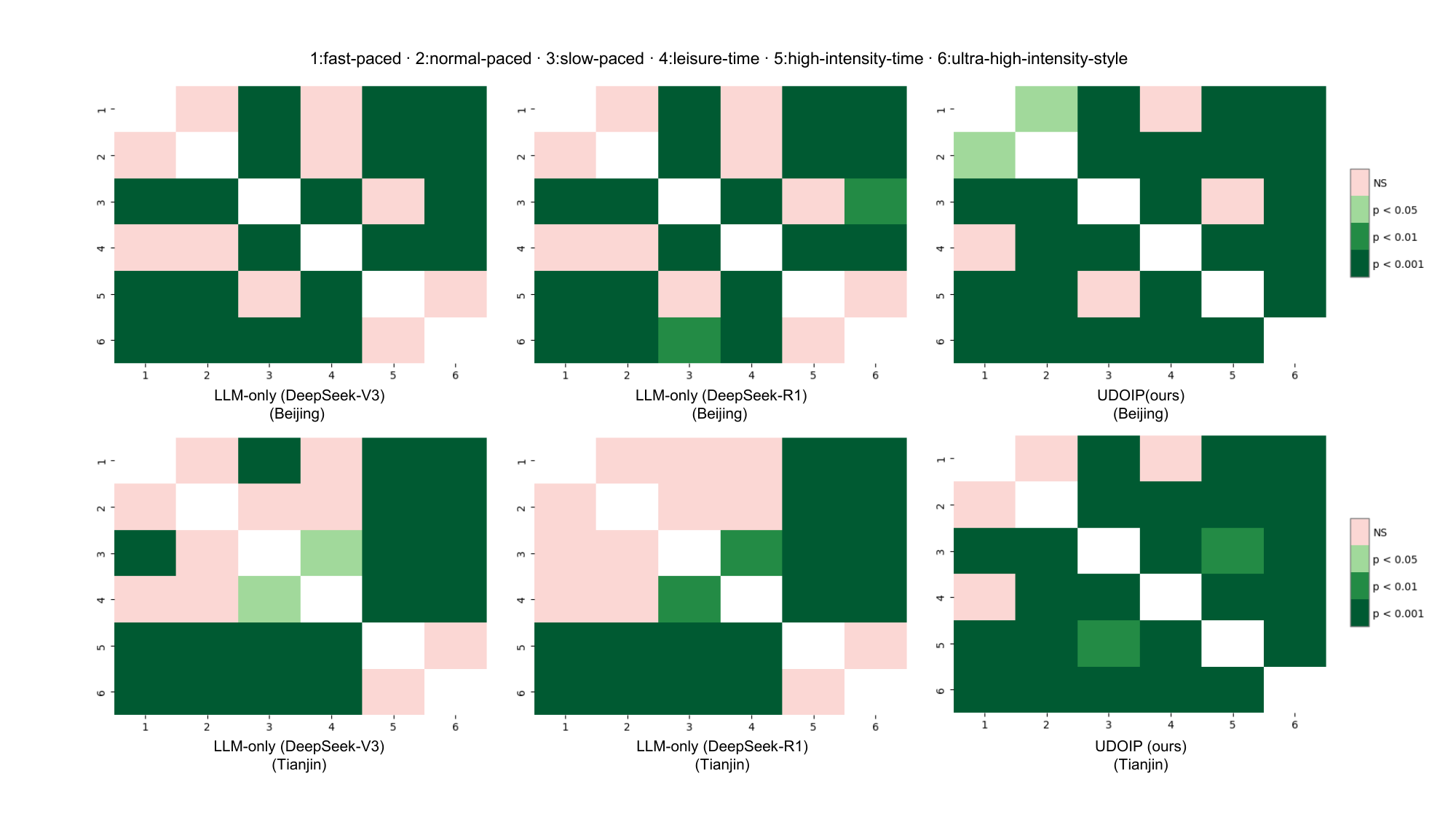}
        \caption{Average Original Recommendation Duration for Two Cities Compared with DeepSeek Model}
        \label{fig:avgd_recommendation_duration}
    \end{minipage}
    \\
    \begin{minipage}[b]{0.45\textwidth}
        \centering
        \includegraphics[width=\textwidth]{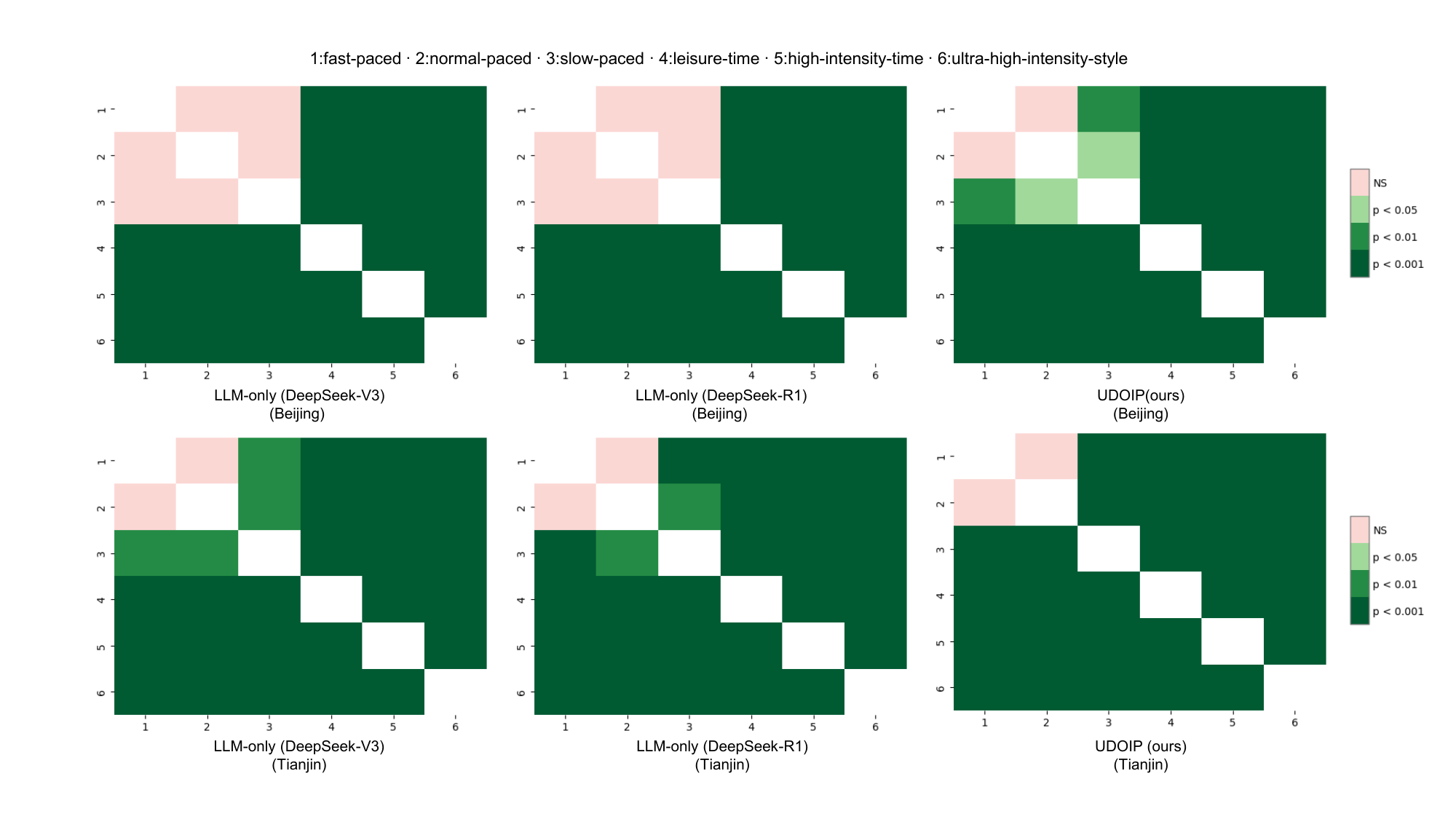}
        \caption{Average Daily Start Time for Two Cities Compared with DeepSeek Model}
        \label{fig:avgd_start_time}
    \end{minipage}
    \hfill
    \begin{minipage}[b]{0.45\textwidth}
        \centering
        \includegraphics[width=\textwidth]{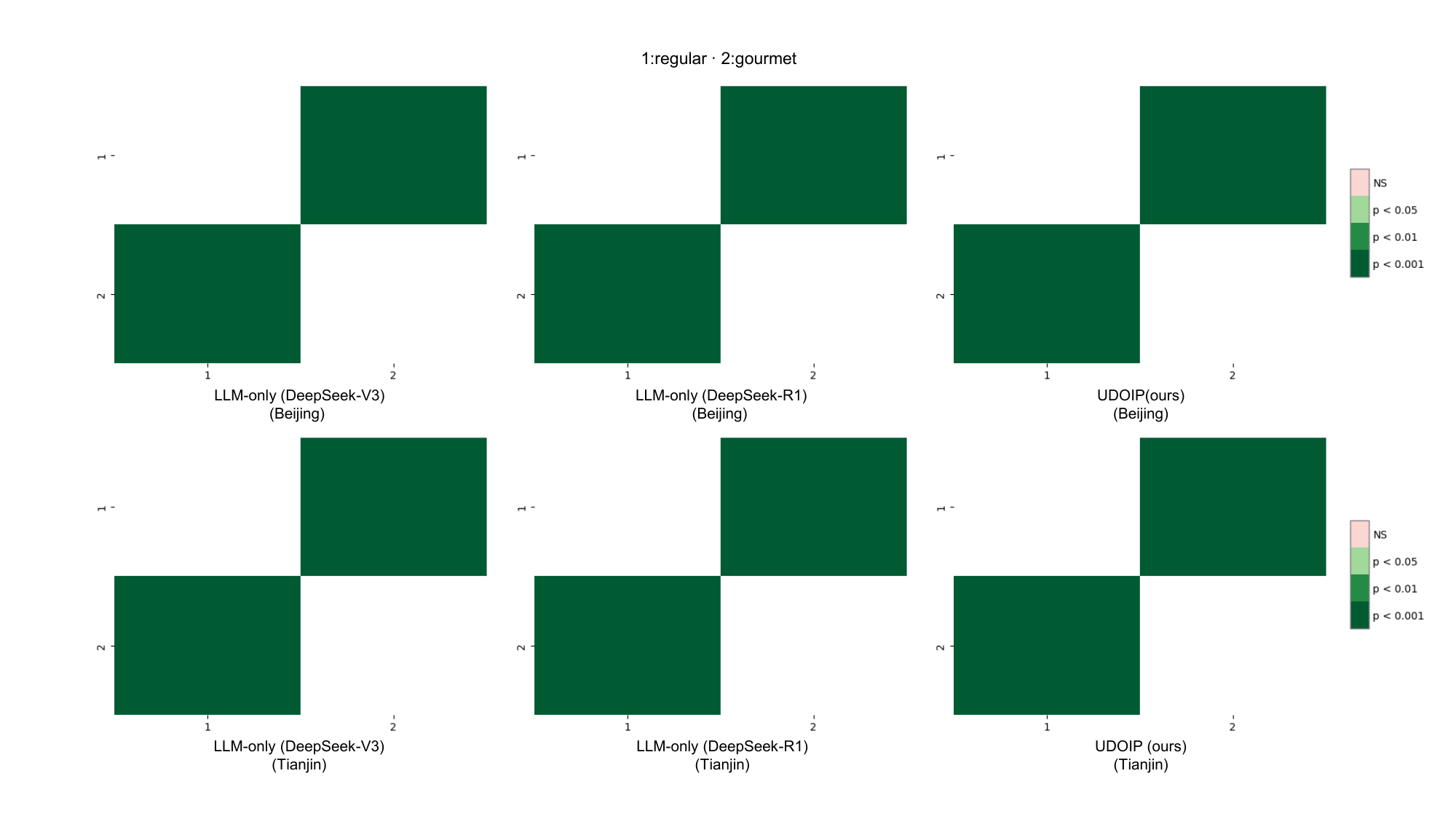}
        \caption{Total Number of Dining Experiences for Two Cities Compared with DeepSeek Model}
        \label{fig:totald_dining_count}
    \end{minipage}
\end{figure*}

In Section~\ref{exp-comparison}, we presented a typical Result of Dunn's test to demonstrate the method's ability to capture preferences across different times of day. To illustrate the ability to capture more diverse preferences, we applied the same analysis to the average daily travel time, the average original recommendation duration, the average daily start time, and the total number of dining experiences for two different cities. The results are presented in Fig.~\ref{fig:avg_daily_travel_time}, Fig.~\ref{fig:avg_recommendation_duration}, Fig.~\ref{fig:avg_start_time}, and Fig.~\ref{fig:total_dining_count}, respectively. The corresponding comparison experiments with the DeepSeek models are presented in Fig.~\ref{fig:avgd_daily_travel_time}, Fig.~\ref{fig:avgd_recommendation_duration}, Fig.~\ref{fig:avgd_start_time}, and Fig.~\ref{fig:totald_dining_count}.

It is evident that for the binary preference of dining, all methods accurately capture it. However, for complex time preferences, our method exhibits stronger discriminative power.

\end{document}